\documentclass{article}

\usepackage{microtype}
\usepackage{graphicx}
\usepackage{subcaption}
\usepackage{booktabs} % for professional tables

\usepackage{hyperref}

\usepackage{amsmath}
\usepackage{amssymb}
\usepackage{mathtools}
\usepackage{amsthm}
\usepackage{pifont}

\usepackage{listings}
\usepackage{tcolorbox}
\tcbuselibrary{breakable, skins} % <--- Add 'skins' here % <--- 关键 1：必须加载 breakable 库
\newtcolorbox{dialogbox}{
  breakable,              % <--- 关键 2：开启跨页参数
  enhanced,               % (可选) 开启增强模式，可以让跨页处的边框处理更平滑
  arc=4mm,
  colback=blue!3,
  colframe=black,
  boxrule=0.5pt,
  fonttitle=\bfseries,
  coltitle=black,
  overlay broken={
  },
}
\usepackage{adjustbox}

\usepackage[preprint]{icml2026}
\usepackage{amsmath}
\usepackage{amssymb}
\usepackage{mathtools}
\usepackage{amsthm}

\usepackage[capitalize,noabbrev]{cleveref}

\theoremstyle{plain}

\theoremstyle{definition}

\theoremstyle{remark}

\usepackage[textsize=tiny]{todonotes}

\usepackage{tcolorbox}
\tcbuselibrary{skins, breakable}
\usepackage{array}

\newtcolorbox{promptbox}[2][]{%
    enhanced,
    breakable,
    colback=gray!5,      % 内容背景色：极浅的灰色
    colframe=black!70,   % 边框颜色：深灰色
    coltitle=white,      % 标题文字颜色
    title={\textbf{#2}}, % 标题加粗
    boxrule=0.8pt,       % 边框宽度
    arc=2pt,             % 圆角半径
    left=3pt, right=3pt, top=3pt, bottom=3pt, % 内边距
    fonttitle=\bfseries\small, % 标题字体
    fontupper=\small,    % 内容字体
    #1
}

\begin{document}

\twocolumn[
  \icmltitle{READY: Reward Discovery for Meta-Black-Box Optimization}



  \icmlsetsymbol{equal}{*}

  \begin{icmlauthorlist}
    \icmlauthor{Zechuan Huang}{scut}
    \icmlauthor{Zhiguang Cao}{smu}
    \icmlauthor{Hongshu Guo}{scut}
    \icmlauthor{Yue-Jiao Gong}{scut}
    \icmlauthor{Zeyuan Ma}{scut}
  \end{icmlauthorlist}

  \icmlaffiliation{scut}{School of Computer Science and Engineering, South China University of Technology, Guangzhou, China}
  \icmlaffiliation{smu}{School of Computing and Information Systems, Singapore Management University, Singapore, Singapore}

  \icmlcorrespondingauthor{Zeyuan Ma}{scut.crazynicolas@gmail.com }

  \icmlkeywords{Machine Learning, ICML}

  \vskip 0.3in
]



\printAffiliationsAndNotice{}  % no special notice (required even if empty)

\begin{abstract}
Meta-Black-Box Optimization (MetaBBO) is an emerging avenue within Optimization community, where algorithm design policy could be meta-learned by reinforcement learning to enhance optimization performance. So far, the reward functions in existing MetaBBO works are designed by human experts, introducing certain design bias and risks of reward hacking. In this paper, we use Large Language Model~(LLM) as an automated reward discovery tool for MetaBBO. Specifically, we consider both effectiveness and efficiency sides. On effectiveness side, we borrow the idea of evolution of heuristics, introducing tailored evolution paradigm in the iterative LLM-based program search process, which ensures continuous improvement. On efficiency side, we additionally introduce multi-task evolution architecture to support parallel reward discovery for diverse MetaBBO approaches. Such parallel process also benefits from knowledge sharing across tasks to accelerate convergence. Empirical results demonstrate that the reward functions discovered by our approach could be helpful for boosting existing MetaBBO works, underscoring the importance of reward design in MetaBBO. We provide READY's project at \url{https://anonymous.4open.science/r/ICML_READY-747F}.
\end{abstract}

\section{Introduction}
% Black-box optimization (BBO) is fundamentally challenging due to its agnostic formulation, where neither the objective function's structure nor its gradients are accessible, requiring effective tactics such as Evolutionary Computation (EC) to navigate complex search spaces. Despite the success of various hand-crafted EC methods in solving diverse BBO problems, the dilemma of limited generalization across distinct tasks persists.The heavy reliance on expertise for task-specific adaptation impede these methods from achieving optimal performance in broader applications.
Meta-Black-Box Optimization (MetaBBO) has recently emerged as a transformative paradigm for solving complex optimization problems where gradient information is unavailable.\cite{yi2023automated,ma2025toward,zhao2023automated,yang2025meta} It adopts a bi-level meta-learning paradigm, where a neural network-based policy(e.g.,RL~\citep{sutton1998reinforcement}) is maintained at the meta-level instructs a BBO optimizer to optimize the target problem. The reward function translates the optimizer's environmental interactions into feedback signals. Since this signal directly influences the performance and generalizability of RL~\cite{lu2025discovery,oh2025discovering}, the design of reward is of great significance. However, with a complex bi-level structure and an uncertain optimization landscape, MetaBBO poses significant challenges for designing effective rewards. Current MetaBBO methods mainly rely on human expertise to craft simple heuristics, and few studies have investigated the efficacy and rationality of reward functions~\cite{nguyen2025importance}.

% motivation(需求)
% user:
This manual dependency severely impedes the scalability and advancement of MetaBBO. 
Developing new architectures necessitates laboriously crafting rewards, typically by adapting insights from existing methods. Moreover, adapting models to diverse real-world scenarios requires manual tuning to align rewards with specific problem objectives.
Furthermore, without fully unlocking the potential of each method via optimized reward functions, fair performance comparisons become nearly impossible. Current hand-crafted rewards are often far from optimal, which obscures the true capabilities of various MetaBBO architectures and complicates further fine-tuning. 
Consequently, we anticipate the development of an automated reward discovery framework for MetaBBO.
% % metabbo field development:
% Without fully inspiring the potential of MetaBBO, it's hard for us to compare different MetaBBO performance. Current reward of MetaBBO is far from optimized, which boder the comparision and further fintuned. We need a automated reward discovery to maximize its performance to compare.
% anticipate an automated framework of reward discovery in MetaBBO

% LLM 
Recently, Large Language Models (LLMs) have revolutionized automated program search. Pioneering frameworks such as FunSearch~\citep{romera2024mathematical}, EoH~\citep{liu2024evolution}, and AlphaEvolve~\citep{novikov2025alphaevolve} have established a new paradigm where LLMs function as intelligent evolutionary operators to discover verifiable mathematical programs. This success has extended to reward engineering, where Eureka~\cite{ma2023eureka} and CARD ~\cite{sun2025large} demonstrate the potential to generate human-level rewards for control tasks via simple, end-to-end pipelines.

Following the paradigm, we propose READY, a framework where LLMs autonomously synthesize ready-to-deploy reward functions for diverse MetaBBO. Specifically, we propose following designs: 1) \textbf{Multitask Program Evolution}: to address the development bottlenecks we mentioned above, we introduce a multitask evolution paradigm into existing program evolution methods. We deploy multiple reward program populations for multiple MetaBBO tasks, leverage their shared methodology insights and architecture to accelerate the search progress; 2) \textbf{Fine-grained Evolution Operators}: we propose five reflection-based code-level evolutionary operators to assure diverse program search behavior; 3) \textbf{Knowledge Transfer}: an explicit knowledge transfer scheme is used to enhance knowledge sharing across different MetaBBO tasks. By using such multitask paradigm, READY is capable of automatically designing desired reward functions for diverse MetaBBO tasks, and more importantly, advancing the performance frontier of MetaBBOs. Empirical results demonstrate that: 1) READY consistently improves diverse MetaBBO's optimization performances, showing advantage against not only their original reward designs but also up-to-date baselines such as Eureka, EoH and ReEvo. 2) READY provides clear and interpretable design insights that co-evolve with the reward program; 3) Surprisingly, we found reward functions designed by READY for a given MetaBBO method could be directly adapted to boost unseen MetaBBO method. Our key contributions are summarized as follows:
\begin{itemize}
    % \item \textbf{First Exploration of Reward Design in MetaBBO:}  We provide a pioneering discussion on the critical role of reward functions in MetaBBO. To the best of our knowledge, this work is the first to systematically explore and automate reward discovery for MetaBBO, addressing a long-standing gap in the field.
    \item We propose READY, which is not only a very first exploration on LLM-based reward design field, but also the first framework to enhance MetaBBO's performance by automated reward design.
    % \item \textbf{The READY Framework:} We propose READY, a niche-based multi-task architecture that extends the LLM-based Automated Design paradigm to reward engineering. With fine-grained evolutionary operators, READY enables LLMs to autonomously discover interpretable and high-quality reward. 
    \item READY is composed by several key designs: 1) multitask paradigm to facilitate knowledge sharing and search effectiveness; 2) fine-grained evolution operators to assure searching diversity; 3) explicit knowledge transfer  scheme to accelerate convergence. 
    % \item \textbf{Superior Performance and Generalizability:} Extensive experiments demonstrate that READY outperforms handcrafted and automated baselines, achieving cost reductions of up to 70\%. The discovered rewards exhibit strong zero-shot generalizability, effectively boosting the performance of unseen MetaBBO.
    \item Experimental results validate that READY shows superior design capability to representative baselines with interpretable design insights. Surprisingly, the searched reward designs could be generalized to unseen MetaBBO tasks.
\end{itemize}

\section{Related Works}
\label{sec:related_works}

\subsection{MetaBBO}
\label{subsec:related_metabbo}
\begin{figure}
    \centering
    \includegraphics[width=0.9\linewidth]{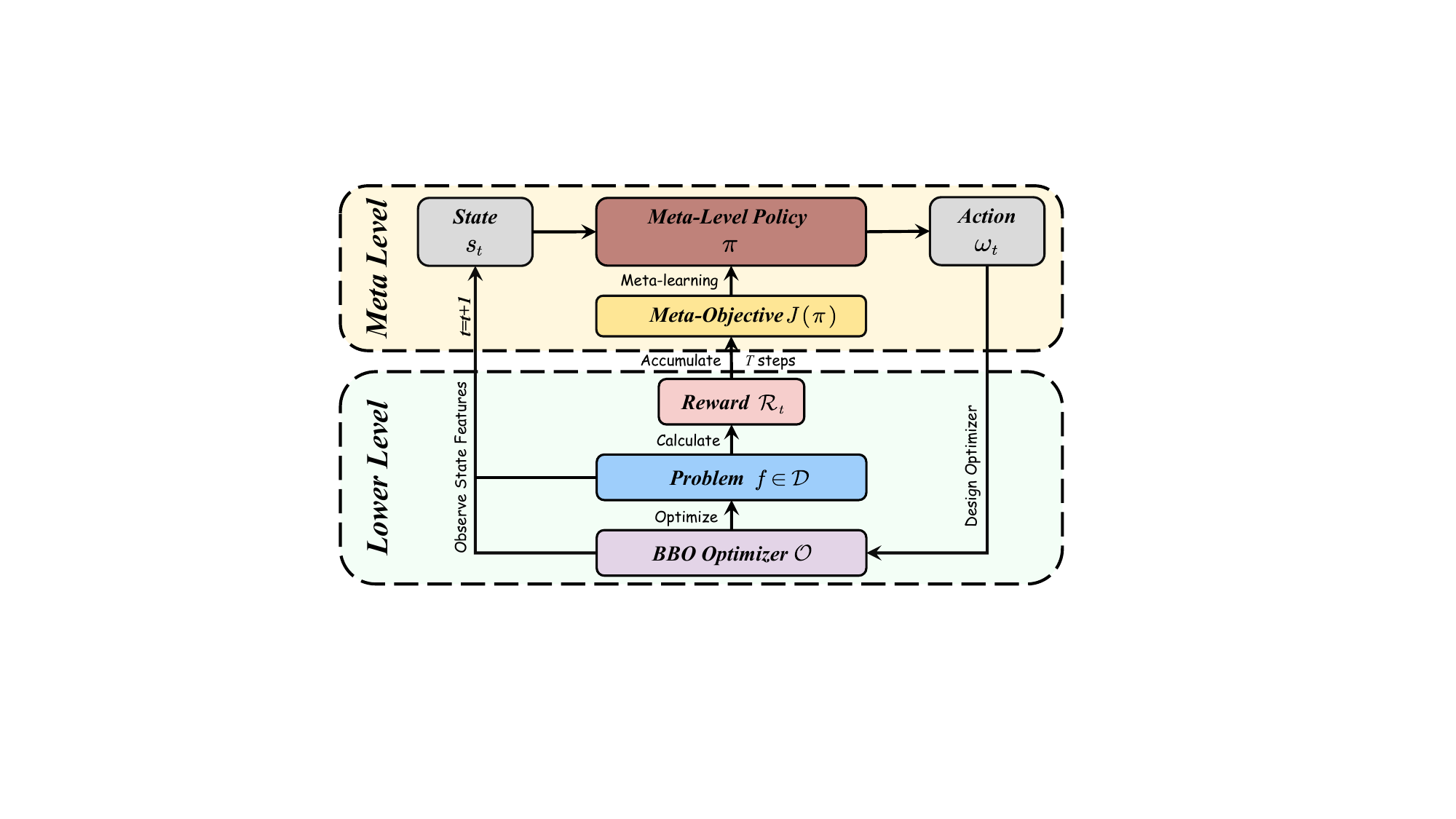}
    \caption{General workflow of MetaBBO approaches.}
    \vspace{-3mm}
    \label{fig:metabbo}
\end{figure}

Meta-Black-Box Optimization~(MetaBBO) aims to automating the design in BBO algorithms by learning-driven paradigm~\cite{ma2025toward,li2024bridging,yang2025meta,ml-metabbo-survey}. As illustrated in Figure~\ref{fig:metabbo}, an algorithm design task $\mathcal{T} = \{\mathcal{D}, \mathcal{O}, \pi\}$ is formalized as a Markov Decision Process (MDP), where $\mathcal{D} = \{\mathcal{D}_\text{train}, \mathcal{D}_\text{test}\}$ is the problem distribution containing a training set and a testing set, $\mathcal{O}$ is the low-level BBO optimizer and $\pi$ is the meta-level policy. Given a problem instance $f \in \mathcal{D}_\text{train}$, for each training optimization step $t$, an optimization state $s_t \in \mathcal{S}$ is extracted from the low-level optimization process. The meta-level policy $\pi$ then outputs algorithm design $\omega_t \in \Omega$ as the action according to $s_t$. $\mathcal{O}$ optimizes $f$ by $\omega^t$ for one step. A reward function $\mathcal{R}$ is used to evaluate the performance gain obtained by this algorithm design action. The training objective is to maximize the expected accumulated reward:
\begin{equation}
J(\pi) = \mathbb{E}_{f \sim \mathcal{D}_\text{train}} \left[ \sum_{t=0}^{T} \gamma^t \mathcal{R}(s_t, \omega_t, f) \right].
\end{equation}

The performance score is commonly evaluated on $\mathcal{D}_\text{test}$ by a normalized objective indicator $\mathcal{F}$ which is computed as:
\begin{equation}\label{eq:evaluate}
    \mathcal{F} = \frac{1}{|\mathcal{D}_\text{test}|} \sum_{i=1}^{|\mathcal{D}_\text{test}|} \left( \mathop{\text{median}}_{j=1}^{\Gamma} \frac{y_{i,j}^{(G)} - y_i^*}{y_{i,j}^{(0)} - y_i^*} \right),
\end{equation}
where $\Gamma$ is the number of test runs, $y_{i,j}^{(0)}$ and $y_{i,j}^{(G)}$ denote the initial and final objective values of the $j$-th run on instance $f_i\in\mathcal{D}_\text{test}$. A smaller $\mathcal{F}$ is better.

% the achieved objective values obtained from applying the trained meta-policy and the low-level optimizer to the test suite $\mathcal{D}_\text{test}$. However, simple averaging of raw objective values is invalid as the problems in $\mathcal{D}_{test}$ exhibit vastly different scales. To address this, a Normalized Fitness Metric $\mathcal{F}$ 

% is computed. Let $y_{i,j}^{(0)}$ and $y_{i,j}^{(G)}$ denote the initial and final objective values of the $j$-th run on instance $f_i\in\mathcal{D}_\text{test}$, and $y_i^*$ be the theoretical optimum. The fitness is derived by normalizing the gap across all test instances:

% where $\Gamma$ is the number of test runs. 
% The inner term represents the optimality gap, where 0 indicates convergence to the optimum and 1 implies no improvement.

The algorithm design tasks MetaBBO could address is quite broad, ranging from operator/algorithm selection/configuration~\cite{guo2024deep,guo2025configx,deddqn,lde} to algorithm generation~\cite{guo2025designxhumancompetitivealgorithmdesigner,chen2024symbolgeneratingflexibleblackbox,zhao2024automated,ma2026llamoco}. Considering the target optimization problem categories, MetaBBO has been instantiated to single-objective~\cite{tan2021differential,han2025enhancing}, multimodal~\cite{lian2024rlemmo}, multi-objective~\cite{tian2025universal,zhang2026reinforcement}, constrained~\cite{li2026reinforcement,hu2023deep} and expensive~\cite{shao2025deep,yao2025fomemo} optimization. Given its swift development speed, automated reward design becomes more and more important.

% In MetaBBO, the algorithm design task is formalized as a Markov Decision Process (MDP), where the environment is the low-level optimization process for a problem instance $f \in \mathcal{D}$ ~\cite{ma2025toward, ma2023metabox, chen2022learning}. The meta-policy $\pi_\theta$ observes the optimization state $s_t \in \mathcal{S}$ and selects an algorithm design action $\omega_t \in \Omega$ to maximize the expected accumulated reward:
% \begin{equation}
% J(\pi_\theta) = \mathbb{E}{f \sim \mathcal{D}} \left[ \sum{t=0}^{T} \gamma^t \mathcal{R}(s_t, \omega_t, f) \right]
% \end{equation}
% where $\mathcal{R}$ is the reward function intended to align with a performance metric $\text{perf}(\cdot)$, such as the reduction in objective cost. A misaligned $\mathcal{R}$ prevents the maximization of cumulative rewards from strictly translating into the minimization of the true optimization cost. Current reliance on hand-crafted heuristics exacerbates this challenge, as manual designs often fail to provide informative or interpretable signals across complex landscapes. Consequently, automating the discovery of an optimal $\mathcal{R}^*$ remains a critical yet underexplored problem.

\subsection{Automated Reward Discovery via LLMs}
\label{subsec:llm_reward_design}

Building upon the success of automated program search~\citep{romera2024mathematical, novikov2025alphaevolve,liu2024evolution}, pioneering explorations have begun to extend LLM capabilities to reward discovery. For instance, Eureka~\citep{ma2023eureka} employs an evolutionary search with LLMs to synthesize reward functions by iteratively refining code based on environmental feedback. Similarly, Text2Reward~\citep{xie2023text2reward} utilizes LLMs to decompose complex semantic specifications into executable sub-reward structures, while CARD~\cite{sun2025large} introduces a self-curated feedback loop to align generated rewards with task objectives via automated dynamic feedback.
Following them, multiples works on the reward discovery emerge.~\cite{lilares,R_star,heng2025boostinguniversalllmreward,su2026end,Hazra2024REvolveRE,yao2025rewardevolutiongraphofthoughts,reward4starcraft,cardenoso2025leveragingllmsrewardfunction,wei2025lerollmdrivenevolutionaryframework}. Unlike classic RL scenarios, MetaBBO shows complexity due to its bi-level framework and uncertainty in the low-level optimization. These challenges motivates this work.

\begin{algorithm}[t]
\caption{READY Overall Optimization Procedure}
\label{alg:ready}
\begin{algorithmic}[1]
\STATE \textbf{Input:} Set of tasks $\{\mathcal{T}_1, \dots, \mathcal{T}_K\}$, Max generations $G_{max}$, Niche size $N$
\STATE \textbf{Output:} Optimal reward set $\mathcal{I}^* = \{I^*_1, \dots, I^*_K\}$
\STATE \textcolor{gray}{// Initialization:} 
\STATE Initialize niches $\{\mathbf{P}_1, \dots, \mathbf{P}_K\}$ (Section~\ref{subsec:init}).
\FOR{generation $g = 1$ to $G_\text{max}$}
    \FOR{each niche $\mathbf{P}_k \in \{\mathbf{P}_1, \dots, \mathbf{P}_K\}$}
        \STATE \textcolor{gray}{// Step 1: Offspring Generation}
        \STATE Generate offspring set $\mathbf{O}_k$ (Section~\ref{subsec:operators}).
        \STATE \textcolor{gray}{// Step 2: Evaluation}
        \FOR{each offspring $I \in \mathbf{O}_k$}
            \STATE Calculate normalized fitness $\mathcal{F}_I$ after train-test.
        \ENDFOR
        \STATE \textcolor{gray}{// Step 3: Survival Selection}
        \STATE Update $\mathbf{P}_k$ by selecting $N$ individuals from $\mathbf{O}_k$ (Section~\ref{subsec:Management}).
    \ENDFOR
    \STATE \textcolor{gray}{// Step 4: Knowledge Transfer}
    \STATE Analyze and decide source and target niches.
    \STATE Generate the transferred individual to replace the worst reward of the target niche (Section~\ref{subsec:knowledge_transfer}).
\ENDFOR
\end{algorithmic}
\end{algorithm}

\section{Methodology}
\label{sec:methodology}
% Overview
% In this section, we first formulate the multitask reward design problem in Section~\ref{subsec:motivation} and then detail the framework designs including the reward function population initialization, reproduction, selection and knowledge transfer in the following Sections~\ref{subsec:init}$\sim$\ref{subsec:knowledge_transfer}.

\subsection{Problem Formulation}
\label{subsec:motivation}
% As introduced in Section \ref{subsec:related_metabbo}, our primary objective is to discover an optimal reward function $r^* \in \Omega$ that minimizes the objective of the MetaBBO-RL, which objective is formulated as finding a set of specialized rewards $\{r^*_k\}_{k=1}^K$ in the multi-task setting. 

Suppose you are a student or researcher who study MetaBBO, there are two design aspects related to reward design you may consider: 1) refer to reward designs in existing MetaBBO methods and redesign proper one for your MetaBBO method; 2) find out which MetaBBO on earth is the nest-performing one by boosting them with optimal reward setting. In fact, both aspects could be formulated as multitask reward discovery problems. Given a set of algorithm design tasks $\{\mathcal{T}_k\}_{k=1}^K$, for each task $\mathcal{T}_k = (\mathcal{D}_k, \mathcal{O}_k, \pi_k)$, the optimization finds the reward $\mathcal{R}$ that minimizes the performance cost $\mathcal{J}$ on each task $\mathcal{T}_k$:
\begin{equation}
\label{READY aim}
    \mathcal{R}^*_k = \mathop{\arg\min}_{\mathcal{R} \in \Omega} \mathcal{F}(\mathcal{R} | \mathcal{T}_k), \quad (k=1 \dots K).
\end{equation}
Here, $\Omega$ is the discrete sourcecodes space of reward functions and 
$\mathcal{F}(\mathcal{R} | \mathcal{T}_k)$ denotes the normalized performance score we defined in Eq.~(\ref{eq:evaluate}).

Searching in discrete space $\Omega$ is intractable. LLMs have inherent advantage in code-level generation and reasoning~\cite{yang2025qwen3}. Inspired by this, our READY leverages LLM-based program evolution to automatically search feasible or even optimal reward design for MetaBBO tasks, with the minimal expertise and labor input required. We present the overall workflow of READY in Algorithm~\ref{alg:ready}, where a multi-task niche-based architecture is deployed to solve Eq.~(\ref{READY aim}). We initialize a separate population~(niche) for each MetaBBO task (Line 4), and let these populations go though intra-population evolution and inter-population knowledge transfer (Line 8). Through iterative interplay with LLMs by these searching operations, READY reflectively propose, analyze and refine reward designs for each MetaBBO tasks. We detail specific designs in READY in the following sections. 
% After initialization of each niche,  As outlined in Algorithm~\ref{alg:ready}, it leverages LLMs to drive evolutionary search and prepare reward functions using their coding ability and RL reward design knowledge.
% Intra-Niche Micro-Evolution.
% Inter-Niche Macro-Collaboration.

\subsection{Initialization with Greedy Sampling}
\label{subsec:init}
% READY integrates the following core components:
% Before detailing the evolutionary process, we formally define the core components that constitute the READY framework:
\subsubsection{Niche Specific Metadata Modeling}
% \subsection{Task Specification and Context Modeling.}
To let the general LLMs be aware of the context of our reward function search tasks, we have to prepare structured information that reflects both the algorithmic and programming context~(termed as metadata in this paper) of each MetaBBO task $\mathcal{T}_k$, just like the prior works~\cite{heng2025boostinguniversalllmreward}. Specifically, we prepare following two kind of context information:

% guide the evolutionary process for the target tasks $\{\mathcal{T}_k\}_{k=1}^K$ defined in Section \ref{subsec:motivation}, we need to establish a structured context~\cite{heng2025boostinguniversalllmreward}. Since the LLM cannot directly perceive the black-box optimization landscape, we introduce Automated Context Extraction to bridge this gap. This mechanism synthesizes the niche specific metadata $\mathcal{M}_k$ through two automated steps:

\textbf{Algorithm Concept:} We feed LLMs with the original research paper of the MetaBBO task $\mathcal{T}_k$, and prompt LLMs for summarization of its major motivations, architectures, algorithmic designs. The summarized metadata provides concise description on $\mathcal{T}_k$ and is termed as $C_{alg}$.

% By synthesizing both the academic paper and source code, the LLM generates a structural description capturing the intent and mechanisms of the target method. This ensures the generated rewards are aligned with the algorithm's operational architecture.
    % \item \textbf{Standardized Hyperparameter Interface:} To define a semantic search space, an LLM analyzer aggregates accessible variables into a unified dictionary. This includes both \textbf{Optimizer States} and \textbf{Agent States}. By exposing upper learning states to the LLM, the LLM can craft sophisticated, state-aware rewards.
    
\textbf{Programming Interface:} To prevent the LLMs generate invalid reward functions due to hallucination, and more importantly, to let the LLMs fully aware of the sourcecodes structure that could be used to design a valid reward function, we feed them with $C_{alg}$ and the implementation sourcecodes of $\mathcal{T}_k$ to analyze usable what variables, functions, classes, storing them into a unified dictionary. We trem this metadata as $C_{code}$. Finally, the overall metadata $\mathcal{M}_k$ is formulated as the union of $C_{alg}$ and $C_{code}$, which will be frequently used in the subsequent evolutionary/knowledge transfer operators, which helps the designed reward aligns with both the algorithmic and code aspects of $\mathcal{T}_k$.

\subsubsection{Gene Encoding}
To harness the reasoning capabilities of LLMs, we decouple the code generation process like EoH ~\citep{liu2024evolution}. Each individual is represented as a pair $I = \langle \textit{Thought}, \textit{Code} \rangle$, where \textit{Thought} provides the natural language description of the reward, summarized by the LLM. \textit{Code} is the reward function code.

\subsubsection{Greedy Niche-based Population Sampling}
\label{subsec:initialization}
Each niche $\mathbf{P}_k$ has $N$ individuals and is explicitly bound to a corresponding metadata $\mathcal{M}_k$. 
Population initialization is conducted independently for each niche following niche specific metadata. Instead of random sampling, this process constructs the population sequentially through three integrated steps to ensure robust starting points:
\begin{itemize}
    \item \textbf{Expert Anchoring:} Each niche is explicitly initialized with the human-designed reward individual $I_{expert}$ deployed to the original MetaBBO method ensuring that the optimization process begins with a proven solution.
    \item \textbf{Iterative In-Context Generation:} To fill the remaining $N-1$ slots, we utilize an LLM to iteratively generate new candidates. This generation is conditioned on the thoughts of all previously accepted individuals within the niche. By explicitly prompting the LLM to generate different logic from existing solutions, we enforce diversity within the initial population.
    \item \textbf{Performance-Based Rejection Sampling:} We apply a strict rejection sampling filter during this expansion. A newly generated individual is first evaluated following Eq.~(\ref{eq:evaluate}). Then it will be added to the niche only if its fitness $\mathcal{F}$ outperforms the human-designed reward individual which ensures that every individual in the initial population effectively dominates the baseline.
\end{itemize}
\begin{promptbox}{Prompt Initialization for the $i$-th Individual:} 
    \textbf{Input:} Niche Specific Metadata $\mathcal{M}$. \\
    \textbf{Context:} Previous Individuals in the Niche \\
    $[ \langle \textit{Thought}_1, \mathcal{F}_1 \rangle, \dots,\langle  \textit{Thought}_{i-1}, \mathcal{F}_{i-1} \rangle ]$. \\
    \textbf{Constraints:} \textit{Diversity:} "The reward should differ from the existing rewards by at least $95\%$." \textit{Quality:} "The reward should perform better than the existing rewards."

    \textbf{Instruction:} Design a new reward based on the context and constraints.
\end{promptbox}

\subsection{Diversified Evolutionary Operators}
\label{subsec:operators}

% READY organizes evolutionary operators into two functional groups. We introduce F-RE and HGE for Intra-Task Refinement, while repurposing social operators (ESE, ECC, SE) for Inter-Task Collaboration to assimilate universal heuristics.
During the intra-niche evolution, READY organizes evolutionary operators into two functional groups: three mutation operators ($M_1\sim M_3$) and two crossover operators ($C_1$ and $C_2$). During the reproduction, for each reward individual, we apply each of these operators one time to generate five offsprings for it.
\begin{itemize}
    % Operator 1: F-RE
    \item \textbf{$M_1$: Local-Reflection Mutation:} 
    Standard evolution often discards poor performers without analysis. Inspired by Eureka~\citep{ma2023eureka}, $M_1$ introduces a distinct "debugging" cycle. First, it identifies the top-$K$ ($K=3$) instances where individual $I$ yields the poorest normalized objective. The LLM analyzes these failure cases to produce a textual reflection, diagnosing why the reward logic misaligned with the agent's policy. Second, this reflection guides the LLM to refine the code, explicitly addressing the identified weaknesses while preserving existing strengths. The detailed prompt could be found in Appendix ~\ref{app:m1prompt}.
    \begin{promptbox}{Prompt: Local-Reflection Mutation}
        \textbf{Reflection Phase:}
        
        \textbf{Input:} Current Code, Top-$K$ Failure Cases + Niche Specific Metadata $\mathcal{M}$ \\
        \textbf{Instruction:} ``Analyze the failure causes on these specific landscapes.''

        \textbf{Mutation Phase:}
        
        \textbf{Input:} Current Code, \textit{Reflection}, Niche Specific Metadata $\mathcal{M}$ \\
        \textbf{Instruction:} ``Incorporate the reflection to modify the code and fix weaknesses.''
    \end{promptbox}

    % Operator 2: HGE
    \item \textbf{$M_2$: History-Reflection Mutation:} 
    Unlike standard operators that only focus on current parents, $M_2$\ exploits the temporal dimension. It retrieves the evolutionary trace within the past $L$ generations (e.g., $L=5$). The LLM acts as a trend analyst, identifying which specific code modifications historically correlated with fitness gains. Based on this observed momentum, it extrapolates the next optimization step.The detailed prompt could be found in Appendix ~\ref{app:m2prompt}.
    \begin{promptbox}{Prompt: History-Reflection Mutation}
        \textbf{Input:} Niche Specific Metadata $\mathcal{M}$ + Current Individual $I_g$. \\
        \textbf{Context:} Evolutionary Trace ($ I_{g-L}, \dots, I_{g-1} $). \\
        \textbf{Instruction:} ``Review the history. Identify which logic changes led to performance gains. Extrapolate this optimization trend to generate the next version.''
    \end{promptbox}
    
    % Operator 5: SE
    \item \textbf{$M_3$: Global-Reflection Mutation:} 
   To capture high-level design patterns, READY maintains a unified global archive $\mathcal{P}_\text{archive}$ containing eliminated rewards from all $K$ niches. The LLM first analyzes and distills global knowledge, abstracting universal design principles to form a summary. This summary then guides the generation of new individuals which will inherit robust methodological patterns.The detailed prompt could be found in Appendix~\ref{app:m3prompt}.
    \begin{promptbox}{Prompt: Global-Reflection Mutation}
        \textbf{Reflection Phase:}
    
        \textbf{Input:} Archive $\mathcal{P}_\text{archive}$. \\
        \textbf{Instruction:} ``Summarize the effective techniques and common patterns in these rewards.'' \\
        \textbf{Mutation Phase:}
        
        \textbf{Input:} Niche Specific Metadata $\mathcal{M}$ + Current Individual $I$ + Archive Summary. \\
        \textbf{Instruction:} ``Use the summary to refine the reward.''
    \end{promptbox}

    % Operator 3: LGE
    \item \textbf{$C_1$: Exploitative Crossover:} 
    $C_1$ implements a social learning strategy that facilitates Implicit Knowledge Transfer. In our multi-task framework, the global best individual $I_{global}^*$ typically originates from a different niche, acting as a carrier of universal meta-heuristics. By prompting the LLM to synthesize the current individual with both the local niche best and the global best, $C_1$ produces a reward that respects local physics while assimilating high-level logic proven successful in other contexts. The detailed prompt could be found in Appendix ~\ref{app:c1prompt}.
    \begin{promptbox}{Prompt: Exploitative Crossover}
        \textbf{Input:} Niche Specific Metadata $\mathcal{M}$ + Current Individual $I$. \\
        \textbf{Context:} Niche Best $I_{niche}^*$ + Global Best $I_{global}^*$. \\
        \textbf{Instruction:} ``Compare Self with the Best solutions. Retain Self's core framework but integrate the superior logic or parameters observed in the Global/Niche Best.''
    \end{promptbox}

    % Operator 4: CN
    \item \textbf{$C_2$: Exploratory Crossover:} 
    $C_2$ functions as a Cross-Task Heuristic Bridge designed to propagate optimization logic across task boundaries. It pairs the current individual $I$ with an elite $I_{partner}^*$ derived from a random distinct niche. The LLM is prompted to retain $I$ as a structural backbone while transplanting unique algorithmic traits from $I_{partner}^*$. This mechanism introduces diversity and prevent overfitting. The detailed prompt could be found in Appendix ~\ref{app:c2prompt}.
    \begin{promptbox}{Prompt: Exploratory Crossover}
        \textbf{Input:} Niche Specific Metadata $\mathcal{M}$ + Current Individual $I$ + Reference Source Individual $I_{partner}^*$. \\
        \textbf{Context:} $I_{partner}^*$ is an elite from a different niche. \\
        \textbf{Instruction:} "Take $I$ as the base structure. Identify unique advantages in $I_{partner}^*$ and inject them into $I$ to create a hybrid solution."
    \end{promptbox}

\end{itemize}
\subsection{Population Management}\label{subsec:Management}
After the reproduction, we have $N$ parents and $5N$ offsprings. To select $N$ elites from the $6N$ individuals for the next generation, we employ following selection rule to compute selection probability $P_i$ for the $i$-th individual:
\begin{equation}
    w_i = \frac{1}{rank_i + 6N}, \quad P_i = \frac{w_i}{\sum_{j=0}^{6N-1} w_j}
\end{equation}
where $rank_i$ is the sorting rank of $i$-th individual in terms of their normalized performance score $F(\cdot|\cdot)$~(ascend). The selection rule helps avoid premature since the selection weights are rescaled.

% a Softened Rank-Based Probability Selection~\cite{binh2024hsevoelevatingautomatic} to maintain diversity and prevent premature convergence in READY. In each generation, we combine the parent population and offspring into a unified pool. All candidates are sorted by fitness (where $rank=0$ denotes the best), and the selection probability $P(rank)$ is calculated as:

% where $6N$ is the total size of the unified pool ($N$ parents generate $5N$ offspring). This distribution provides a survival advantage to elite individuals while explicitly preserving promising sub-optimal solutions to sustain genetic diversity.

\subsection{Knowledge Transfer}
\label{subsec:knowledge_transfer}

% To facilitate explicit collaboration across different optimization tasks, READY introduces a \textbf{Knowledge Transfer (KT)} operator that functions as a high-level meta-controller. This process is divided into two distinct phases:
Recent research works reveal that LLMs may naturally be a multitask solver~\cite{ong2024llm2feadiscovernovel,wu2024advancingautomatedknowledge}.
A key insight is that: for the reward design task we consider in this paper, the similarity~(methodology, architecture, learning methods, etc.) between two MetaBBO tasks is a useful information for them to promote designed reward not only by intra-niche evolution, but also inter-niche knowledge transfer. This motivates us to introduce Knowledge Transfer (KT) operator, which operates by first identifying~(analyzing) similar task pairs and then suggesting how to transfer the reward code of source tasks to target tasks. Specifically, the LLM receives a global view of the population and a transfer history log $H$. The global view includes the niche metadata ($\mathcal{M}_{1 \dots K}$) and thoughts of the current individuals in each niche. The transfer history log records the structural details of past transfers, specifically the reflection behind the transfer, the transfer strategy, and the result of the transfer.
Based on the context, the LLM identifies $K$ pathways (Source $\to$ Target) and generates the corresponding reflection and strategy. The detailed prompt is in Appendix~\ref{app:ktprompt}.
% To facilitate explicit knowledge sharing across different, READY introduces a Knowledge Transfer (KT) operator that functions as a high-level meta-controller. Via a dynamic transfer history log, a KT implements a self-reflective learning mechanism, allowing the system to progressively learn the global task topology by identifying high-yield transfer pathways.

% \paragraph{Phase 1: Meta-Controller Decision.}
% The LLM receives details of past transfers (the 'why', 'how', and 'results') and a \textbf{Global View} of the population which includes the task specifications ($\mathcal{S}_{1 \dots K}$) and the design rationales \textit{Thoughts} of the current elites in each niche. Here, a transfer event is recorded as Success if the transferred individual surpasses the performance of the worst individual in the target niche.
% The LLM receives a global view of the population and a transfer history log $H$. The global view includes the niche specific metadata ($\mathcal{M}_{1 \dots K}$) and thoughts of the current individuals in each niche. The transfer history log records the structural details of past transfers, specifically the reflection behind the transfer, the transfer strategy, and the result of the transfer.
% Based on the context, the LLM identifies $K$ pathways (Source $\to$ Target) and generates the corresponding reflection and strategy. The detailed prompt is in Appendix ~\ref{app:ktprompt}.

\begin{promptbox}{Prompt: KT - Reflection Phase}
    \textbf{Input:} Metadata $\{\mathcal{M}_k\}_{k=1}^K$ 
    Individuals \textit{Thoughts} from all niches + Transfer History $H$.
    
    \textbf{Instruction:} ``Analyze and suggest $K$ source-target pathways with high potential for positive knowledge transfer. For each pathway, provide a reflection and transfer strategy suggestion.''
\end{promptbox}

% \paragraph{Phase 2: Execution and Adaptation.}
The LLM transforms $I_{source}$ into the target context, guided by $\mathcal{M}_{target}$ with the reflection and the strategy generated above. The adapted reward replaces the lowest-ranked individual in the target niche.
\begin{promptbox}{Prompt: KT - Execution}
    \textbf{Input:} Source Code $I_{source}$ + Target Metadata $\mathcal{M}_{target}$ + \textit{Reflection} + Transfer Strategy. \\
    \textbf{Instruction:} ``You are transplanting a reward from the Source to the Target task. Strictly apply the provided strategy to adapt the code logic to the new task constraints defined in $\mathcal{M}_{target}$.''
\end{promptbox}

\begin{table*}[t]
\centering
\caption{Comparison of discovered rewards on test functions. We report the mean minimum cost $\pm$ standard deviation over 51 independent runs (lower is better). \textbf{Bold} indicates the best performance, and \underline{underlined} denotes the second best.  Avg. Rank represents the average ranking of each method across all test functions. Note that READY discovers rewards for all three MetaBBO in a single run, whereas baselines are optimized independently for each task.}
\label{tab:main_res}
\resizebox{\textwidth}{!}{%
\renewcommand{\arraystretch}{1.1}
\begin{tabular}{l|ccccc|ccccc|ccccc}
\hline
& \multicolumn{5}{c|}{DEDQN} 
& \multicolumn{5}{c|}{RLEPSO} 
& \multicolumn{5}{c}{RLDAS} 
\\ \cline{2-16}
& Handcrafted & EoH & ReEvo & Eureka & \textbf{READY} 
& Handcrafted & EoH & ReEvo & Eureka & \textbf{READY} 
& Handcrafted & EoH & ReEvo & Eureka & \textbf{READY} 
\\ \hline

\textit{Attractive Sector}
& \begin{tabular}[c]{@{}c@{}}7.84e+03\\ $\pm$9.62e+03\end{tabular} 
& \begin{tabular}[c]{@{}c@{}}\underline{5.86e+03}\\ $\pm$8.69e+03\end{tabular} 
& \begin{tabular}[c]{@{}c@{}}7.99e+03\\ $\pm$9.86e+03\end{tabular} 
& \begin{tabular}[c]{@{}c@{}}8.04e+03\\ $\pm$9.56e+03\end{tabular} 
& \textbf{\begin{tabular}[c]{@{}c@{}}2.35e+03\\ $\pm$4.21e+03\end{tabular}}
& \begin{tabular}[c]{@{}c@{}}6.17e-03\\ $\pm$1.27e-02\end{tabular} 
& \begin{tabular}[c]{@{}c@{}}1.39e-02\\ $\pm$4.93e-02\end{tabular} 
& \begin{tabular}[c]{@{}c@{}}\underline{4.14e-03}\\ $\pm$2.23e-02\end{tabular}
& \begin{tabular}[c]{@{}c@{}}2.39e-02\\ $\pm$9.38e-02\end{tabular} 
& \textbf{\begin{tabular}[c]{@{}c@{}}4.09e-03\\ $\pm$8.49e-03\end{tabular} }
& \begin{tabular}[c]{@{}c@{}}1.37e-01\\ $\pm$1.59e-01\end{tabular} 
& \begin{tabular}[c]{@{}c@{}}1.44e-01\\ $\pm$1.60e-01\end{tabular} 
& \begin{tabular}[c]{@{}c@{}}1.44e-01\\ $\pm$1.60e-01\end{tabular} 
& \begin{tabular}[c]{@{}c@{}}\underline{1.36e-01}\\ $\pm$1.43e-01\end{tabular} 
& \textbf{\begin{tabular}[c]{@{}c@{}}1.28e-01\\ $\pm$1.40e-01\end{tabular}}
\\

\textit{Bent Cigar}
& \begin{tabular}[c]{@{}c@{}}3.17e+07\\ $\pm$1.03e+07\end{tabular} 
& \begin{tabular}[c]{@{}c@{}}\underline{2.90e+07}\\ $\pm$9.93e+06\end{tabular} 
& \begin{tabular}[c]{@{}c@{}}3.27e+07\\ $\pm$1.13e+07\end{tabular} 
& \begin{tabular}[c]{@{}c@{}}3.15e+07\\ $\pm$9.72e+06\end{tabular} 
& \textbf{\begin{tabular}[c]{@{}c@{}}2.64e+07\\ $\pm$1.01e+07\end{tabular}}
& \begin{tabular}[c]{@{}c@{}}1.45e+00\\ $\pm$1.70e+00\end{tabular} 
& \textbf{\begin{tabular}[c]{@{}c@{}}1.20e+00\\ $\pm$1.43e+00\end{tabular}}
& \begin{tabular}[c]{@{}c@{}}\underline{1.23e+00}\\ $\pm$2.10e+00\end{tabular} 
& \begin{tabular}[c]{@{}c@{}}1.51e+00\\ $\pm$2.00e+00\end{tabular} 
& \begin{tabular}[c]{@{}c@{}}1.43e+00\\ $\pm$2.41e+00\end{tabular} 
& \textbf{\begin{tabular}[c]{@{}c@{}}6.66e+00\\ $\pm$1.88e+01\end{tabular}}
& \begin{tabular}[c]{@{}c@{}}\underline{6.75e+00}\\ $\pm$1.88e+01\end{tabular} 
& \begin{tabular}[c]{@{}c@{}}6.76e+00\\ $\pm$1.88e+01\end{tabular} 
& \begin{tabular}[c]{@{}c@{}}6.87e+00\\ $\pm$1.87e+01\end{tabular} 
& \begin{tabular}[c]{@{}c@{}}6.88e+00\\ $\pm$1.89e+01\end{tabular} 
\\

\textit{Buche Rastrigin}
& \begin{tabular}[c]{@{}c@{}}4.14e+02\\ $\pm$1.10e+02\end{tabular} 
& \begin{tabular}[c]{@{}c@{}}\underline{4.00e+02}\\ $\pm$1.31e+02\end{tabular} 
& \begin{tabular}[c]{@{}c@{}}4.34e+02\\ $\pm$1.44e+02\end{tabular} 
& \begin{tabular}[c]{@{}c@{}}4.61e+02\\ $\pm$1.52e+02\end{tabular} 
& \textbf{\begin{tabular}[c]{@{}c@{}}3.18e+02\\ $\pm$1.07e+02\end{tabular}}
& \begin{tabular}[c]{@{}c@{}}5.75e+01\\ $\pm$2.82e+01\end{tabular} 
& \begin{tabular}[c]{@{}c@{}}\underline{5.27e+01}\\ $\pm$2.18e+01\end{tabular} 
& \begin{tabular}[c]{@{}c@{}}6.54e+01\\ $\pm$3.19e+01\end{tabular} 
& \textbf{\begin{tabular}[c]{@{}c@{}}5.23e+01\\ $\pm$2.03e+01\end{tabular}}
& \begin{tabular}[c]{@{}c@{}}6.52e+01\\ $\pm$2.80e+01\end{tabular} 
& \begin{tabular}[c]{@{}c@{}}3.11e+01\\ $\pm$7.25e+00\end{tabular} 
& \begin{tabular}[c]{@{}c@{}}\underline{3.09e+01}\\ $\pm$6.88e+00\end{tabular}
& \begin{tabular}[c]{@{}c@{}}3.11e+01\\ $\pm$7.07e+00\end{tabular} 
& \begin{tabular}[c]{@{}c@{}}3.12e+01\\ $\pm$6.99e+00\end{tabular} 
& \textbf{\begin{tabular}[c]{@{}c@{}}3.09e+01\\ $\pm$6.87e+00\end{tabular}}
\\

\textit{Composite Grie rosen}
& \begin{tabular}[c]{@{}c@{}}1.23e+01\\ $\pm$2.07e+00\end{tabular} 
& \begin{tabular}[c]{@{}c@{}}1.16e+01\\ $\pm$2.01e+00\end{tabular} 
& \begin{tabular}[c]{@{}c@{}}\underline{1.14e+01}\\ $\pm$2.00e+00\end{tabular} 
& \textbf{\begin{tabular}[c]{@{}c@{}}1.10e+01\\ $\pm$1.80e+00\end{tabular}}
& \begin{tabular}[c]{@{}c@{}}1.19e+01\\ $\pm$2.00e+00\end{tabular} 
& \begin{tabular}[c]{@{}c@{}}1.34e+00\\ $\pm$5.84e-01\end{tabular} 
& \begin{tabular}[c]{@{}c@{}}1.75e+00\\ $\pm$5.20e-01\end{tabular} 
& \begin{tabular}[c]{@{}c@{}}\underline{1.21e+00}\\ $\pm$5.40e-01\end{tabular} 
& \begin{tabular}[c]{@{}c@{}}1.73e+00\\ $\pm$5.07e-01\end{tabular} 
& \textbf{\begin{tabular}[c]{@{}c@{}}1.13e+00\\ $\pm$5.59e-01\end{tabular}}
& \begin{tabular}[c]{@{}c@{}}1.67e+00\\ $\pm$6.97e-01\end{tabular} 
& \begin{tabular}[c]{@{}c@{}}\underline{1.62e+00}\\ $\pm$6.99e-01\end{tabular}
& \begin{tabular}[c]{@{}c@{}}1.62e+00\\ $\pm$7.00e-01\end{tabular}
& \begin{tabular}[c]{@{}c@{}}1.63e+00\\ $\pm$6.99e-01\end{tabular} 
& \textbf{\begin{tabular}[c]{@{}c@{}}1.62e+00\\ $\pm$6.94e-01\end{tabular}}
\\

\textit{Different Powers}
& \begin{tabular}[c]{@{}c@{}}1.22e+01\\ $\pm$2.47e+00\end{tabular} 
& \textbf{\begin{tabular}[c]{@{}c@{}}1.11e+01\\ $\pm$2.81e+00\end{tabular}}
& \begin{tabular}[c]{@{}c@{}}1.17e+01\\ $\pm$2.72e+00\end{tabular} 
& \begin{tabular}[c]{@{}c@{}}1.15e+01\\ $\pm$2.96e+00\end{tabular} 
& \begin{tabular}[c]{@{}c@{}}\underline{1.12e+01}\\ $\pm$3.10e+00\end{tabular} 
& \begin{tabular}[c]{@{}c@{}}2.38e-04\\ $\pm$9.70e-05\end{tabular} 
& \begin{tabular}[c]{@{}c@{}}3.66e-04\\ $\pm$1.47e-04\end{tabular} 
& \begin{tabular}[c]{@{}c@{}}\underline{1.10e-04}\\ $\pm$4.50e-05\end{tabular} 
& \begin{tabular}[c]{@{}c@{}}3.75e-04\\ $\pm$2.10e-04\end{tabular} 
& \textbf{\begin{tabular}[c]{@{}c@{}}1.06e-04\\ $\pm$6.02e-05\end{tabular}}
& \begin{tabular}[c]{@{}c@{}}5.19e-04\\ $\pm$4.29e-04\end{tabular} 
& \textbf{\begin{tabular}[c]{@{}c@{}}5.08e-04\\ $\pm$4.28e-04\end{tabular}}
& \begin{tabular}[c]{@{}c@{}}\underline{5.11e-04}\\ $\pm$4.26e-04\end{tabular} 
& \begin{tabular}[c]{@{}c@{}}5.20e-04\\ $\pm$4.28e-04\end{tabular} 
& \begin{tabular}[c]{@{}c@{}}5.12e-04\\ $\pm$4.22e-04\end{tabular} 
\\

\textit{Discus}
& \begin{tabular}[c]{@{}c@{}}8.34e+02\\ $\pm$1.05e+03\end{tabular} 
& \textbf{\begin{tabular}[c]{@{}c@{}}7.50e+02\\ $\pm$8.12e+02\end{tabular}}
& \begin{tabular}[c]{@{}c@{}}1.17e+03\\ $\pm$1.62e+03\end{tabular} 
& \begin{tabular}[c]{@{}c@{}}8.91e+02\\ $\pm$1.22e+03\end{tabular} 
& \begin{tabular}[c]{@{}c@{}}\underline{8.01e+02}\\ $\pm$1.20e+03\end{tabular} 
& \textbf{\begin{tabular}[c]{@{}c@{}}1.36e+01\\ $\pm$6.88e+00\end{tabular}}
& \begin{tabular}[c]{@{}c@{}}1.42e+01\\ $\pm$6.36e+00\end{tabular} 
& \begin{tabular}[c]{@{}c@{}}1.46e+01\\ $\pm$8.33e+00\end{tabular} 
& \begin{tabular}[c]{@{}c@{}}\underline{1.37e+01}\\ $\pm$8.72e+00\end{tabular} 
& \begin{tabular}[c]{@{}c@{}}1.80e+01\\ $\pm$1.12e+01\end{tabular} 
& \begin{tabular}[c]{@{}c@{}}3.38e+00\\ $\pm$4.49e+00\end{tabular} 
& \begin{tabular}[c]{@{}c@{}}3.26e+00\\ $\pm$4.39e+00\end{tabular} 
& \begin{tabular}[c]{@{}c@{}}3.37e+00\\ $\pm$4.40e+00\end{tabular} 
& \textbf{\begin{tabular}[c]{@{}c@{}}3.12e+00\\ $\pm$4.29e+00\end{tabular}}
& \begin{tabular}[c]{@{}c@{}}\underline{3.20e+00}\\ $\pm$4.37e+00\end{tabular} 
\\

\textit{Ellipsoidal high cond}
& \begin{tabular}[c]{@{}c@{}}3.92e+05\\ $\pm$2.01e+05\end{tabular} 
& \begin{tabular}[c]{@{}c@{}}\underline{3.57e+05}\\ $\pm$1.96e+05\end{tabular} 
& \begin{tabular}[c]{@{}c@{}}3.80e+05\\ $\pm$1.77e+05\end{tabular} 
& \begin{tabular}[c]{@{}c@{}}4.04e+05\\ $\pm$2.19e+05\end{tabular} 
& \textbf{\begin{tabular}[c]{@{}c@{}}2.92e+05\\ $\pm$1.74e+05\end{tabular}}
& \begin{tabular}[c]{@{}c@{}}6.80e+02\\ $\pm$7.10e+02\end{tabular} 
& \begin{tabular}[c]{@{}c@{}}8.15e+02\\ $\pm$6.36e+02\end{tabular} 
& \begin{tabular}[c]{@{}c@{}}\underline{6.53e+02}\\ $\pm$1.21e+03\end{tabular} 
& \begin{tabular}[c]{@{}c@{}}9.21e+02\\ $\pm$9.77e+02\end{tabular} 
& \textbf{\begin{tabular}[c]{@{}c@{}}5.71e+02\\ $\pm$5.80e+02\end{tabular}}
& \begin{tabular}[c]{@{}c@{}}2.81e+02\\ $\pm$4.86e+02\end{tabular} 
& \begin{tabular}[c]{@{}c@{}}\underline{2.79e+02}\\ $\pm$4.80e+02\end{tabular} 
& \begin{tabular}[c]{@{}c@{}}2.82e+02\\ $\pm$4.80e+02\end{tabular} 
& \begin{tabular}[c]{@{}c@{}}2.83e+02\\ $\pm$4.85e+02\end{tabular} 
& \textbf{\begin{tabular}[c]{@{}c@{}}2.67e+02\\ $\pm$4.84e+02\end{tabular}}
\\

\textit{Gallagher 21Peaks}
& \begin{tabular}[c]{@{}c@{}}6.23e+01\\ $\pm$1.01e+01\end{tabular} 
& \begin{tabular}[c]{@{}c@{}}\underline{6.00e+01}\\ $\pm$1.02e+01\end{tabular} 
& \begin{tabular}[c]{@{}c@{}}6.21e+01\\ $\pm$1.09e+01\end{tabular} 
& \begin{tabular}[c]{@{}c@{}}6.22e+01\\ $\pm$1.08e+01\end{tabular} 
& \textbf{\begin{tabular}[c]{@{}c@{}}5.93e+01\\ $\pm$1.22e+01\end{tabular}}
& \begin{tabular}[c]{@{}c@{}}7.97e+00\\ $\pm$1.09e+01\end{tabular} 
& \begin{tabular}[c]{@{}c@{}}\underline{7.82e+00}\\ $\pm$1.07e+01\end{tabular} 
& \begin{tabular}[c]{@{}c@{}}9.26e+00\\ $\pm$1.11e+01\end{tabular} 
& \begin{tabular}[c]{@{}c@{}}7.85e+00\\ $\pm$9.63e+00\end{tabular}
& \textbf{\begin{tabular}[c]{@{}c@{}}7.79e+00\\ $\pm$9.95e+00\end{tabular} }
& \begin{tabular}[c]{@{}c@{}}5.81e-01\\ $\pm$7.94e-01\end{tabular} 
& \begin{tabular}[c]{@{}c@{}}5.39e-01\\ $\pm$7.72e-01\end{tabular} 
& \begin{tabular}[c]{@{}c@{}}5.46e-01\\ $\pm$7.72e-01\end{tabular} 
& \textbf{\begin{tabular}[c]{@{}c@{}}5.02e-01\\ $\pm$7.56e-01\end{tabular}}
& \begin{tabular}[c]{@{}c@{}}\underline{5.30e-01}\\ $\pm$7.75e-01\end{tabular} 
\\

\textit{Katsuura}
& \textbf{\begin{tabular}[c]{@{}c@{}}3.42e+00\\ $\pm$7.80e-01\end{tabular}}
& \begin{tabular}[c]{@{}c@{}}3.82e+00\\ $\pm$8.44e-01\end{tabular} 
& \begin{tabular}[c]{@{}c@{}}\underline{3.51e+00}\\ $\pm$7.89e-01\end{tabular} 
& \begin{tabular}[c]{@{}c@{}}3.60e+00\\ $\pm$7.63e-01\end{tabular} 
& \begin{tabular}[c]{@{}c@{}}3.59e+00\\ $\pm$8.82e-01\end{tabular} 
& \begin{tabular}[c]{@{}c@{}}1.16e+00\\ $\pm$3.78e-01\end{tabular} 
& \begin{tabular}[c]{@{}c@{}}1.33e+00\\ $\pm$3.11e-01\end{tabular} 
& \begin{tabular}[c]{@{}c@{}}\underline{9.81e-01}\\ $\pm$3.66e-01\end{tabular}
& \begin{tabular}[c]{@{}c@{}}1.31e+00\\ $\pm$3.29e-01\end{tabular} 
& \textbf{\begin{tabular}[c]{@{}c@{}}9.74e-01\\ $\pm$3.64e-01\end{tabular} }
& \textbf{\begin{tabular}[c]{@{}c@{}}1.26e+00\\ $\pm$2.63e-01\end{tabular}}
& \begin{tabular}[c]{@{}c@{}}1.29e+00\\ $\pm$2.75e-01\end{tabular} 
& \begin{tabular}[c]{@{}c@{}}1.28e+00\\ $\pm$2.76e-01\end{tabular} 
& \begin{tabular}[c]{@{}c@{}}\underline{1.26e+00}\\ $\pm$2.73e-01\end{tabular}
& \begin{tabular}[c]{@{}c@{}}1.27e+00\\ $\pm$2.63e-01\end{tabular} 
\\

\textit{Lunacek bi Rastrigin}
& \begin{tabular}[c]{@{}c@{}}1.69e+02\\ $\pm$1.90e+01\end{tabular} 
& \begin{tabular}[c]{@{}c@{}}\underline{1.60e+02}\\ $\pm$1.91e+01\end{tabular} 
& \begin{tabular}[c]{@{}c@{}}1.63e+02\\ $\pm$1.89e+01\end{tabular} 
& \textbf{\begin{tabular}[c]{@{}c@{}}1.59e+02\\ $\pm$1.82e+01\end{tabular}}
& \begin{tabular}[c]{@{}c@{}}1.62e+02\\ $\pm$1.95e+01\end{tabular} 
& \begin{tabular}[c]{@{}c@{}}2.55e+01\\ $\pm$7.89e+00\end{tabular} 
& \begin{tabular}[c]{@{}c@{}}2.94e+01\\ $\pm$9.21e+00\end{tabular} 
& \textbf{\begin{tabular}[c]{@{}c@{}}2.43e+01\\ $\pm$6.99e+00\end{tabular}}
& \begin{tabular}[c]{@{}c@{}}2.76e+01\\ $\pm$9.19e+00\end{tabular} 
& \begin{tabular}[c]{@{}c@{}}\underline{2.50e+01}\\ $\pm$7.38e+00\end{tabular} 
& \begin{tabular}[c]{@{}c@{}}\underline{3.96e+01}\\ $\pm$7.93e+00\end{tabular} 
& \begin{tabular}[c]{@{}c@{}}3.98e+01\\ $\pm$8.15e+00\end{tabular} 
& \begin{tabular}[c]{@{}c@{}}4.00e+01\\ $\pm$7.98e+00\end{tabular} 
& \textbf{\begin{tabular}[c]{@{}c@{}}3.95e+01\\ $\pm$8.31e+00\end{tabular}}
& \begin{tabular}[c]{@{}c@{}}3.99e+01\\ $\pm$8.06e+00\end{tabular} 
\\

\textit{Rosenbrock original}
& \begin{tabular}[c]{@{}c@{}}1.39e+04\\ $\pm$6.73e+03\end{tabular} 
& \begin{tabular}[c]{@{}c@{}}\underline{1.24e+04}\\ $\pm$6.12e+03\end{tabular} 
& \begin{tabular}[c]{@{}c@{}}1.35e+04\\ $\pm$6.54e+03\end{tabular} 
& \begin{tabular}[c]{@{}c@{}}1.36e+04\\ $\pm$6.07e+03\end{tabular} 
& \textbf{\begin{tabular}[c]{@{}c@{}}1.13e+04\\ $\pm$5.74e+03\end{tabular}}
& \begin{tabular}[c]{@{}c@{}}3.17e+00\\ $\pm$1.66e+00\end{tabular} 
& \begin{tabular}[c]{@{}c@{}}3.78e+00\\ $\pm$1.85e+00\end{tabular} 
& \begin{tabular}[c]{@{}c@{}}\underline{3.10e+00}\\ $\pm$1.66e+00\end{tabular}
& \begin{tabular}[c]{@{}c@{}}3.79e+00\\ $\pm$1.88e+00\end{tabular} 
& \textbf{\begin{tabular}[c]{@{}c@{}}2.95e+00\\ $\pm$2.17e+00\end{tabular} }
& \begin{tabular}[c]{@{}c@{}}2.70e+00\\ $\pm$1.56e+00\end{tabular} 
& \begin{tabular}[c]{@{}c@{}}2.65e+00\\ $\pm$1.48e+00\end{tabular} 
& \begin{tabular}[c]{@{}c@{}}2.65e+00\\ $\pm$1.50e+00\end{tabular} 
& \begin{tabular}[c]{@{}c@{}}\underline{2.64e+00}\\ $\pm$1.53e+00\end{tabular} 
& \textbf{\begin{tabular}[c]{@{}c@{}}2.60e+00\\ $\pm$1.54e+00\end{tabular}}
\\

\textit{Rosenbrock rotated}
& \begin{tabular}[c]{@{}c@{}}1.22e+04\\ $\pm$5.87e+03\end{tabular} 
& \begin{tabular}[c]{@{}c@{}}1.08e+04\\ $\pm$4.93e+03\end{tabular} 
& \begin{tabular}[c]{@{}c@{}}\underline{1.07e+04}\\ $\pm$6.59e+03\end{tabular} 
& \textbf{\begin{tabular}[c]{@{}c@{}}9.47e+03\\ $\pm$5.01e+03\end{tabular}}
& \begin{tabular}[c]{@{}c@{}}1.20e+04\\ $\pm$4.99e+03\end{tabular} 
& \begin{tabular}[c]{@{}c@{}}3.81e+00\\ $\pm$2.04e+00\end{tabular} 
& \begin{tabular}[c]{@{}c@{}}5.96e+00\\ $\pm$8.87e+00\end{tabular} 
& \begin{tabular}[c]{@{}c@{}}\underline{3.26e+00}\\ $\pm$1.93e+00\end{tabular} 
& \begin{tabular}[c]{@{}c@{}}5.65e+00\\ $\pm$9.16e+00\end{tabular} 
& \textbf{\begin{tabular}[c]{@{}c@{}}2.91e+00\\ $\pm$1.50e+00\end{tabular}}
& \begin{tabular}[c]{@{}c@{}}\underline{4.50e+00}\\ $\pm$1.86e+00\end{tabular} 
& \begin{tabular}[c]{@{}c@{}}4.51e+00\\ $\pm$1.86e+00\end{tabular} 
& \begin{tabular}[c]{@{}c@{}}4.51e+00\\ $\pm$1.86e+00\end{tabular} 
& \begin{tabular}[c]{@{}c@{}}4.51e+00\\ $\pm$1.88e+00\end{tabular} 
& \textbf{\begin{tabular}[c]{@{}c@{}}4.48e+00\\ $\pm$1.85e+00\end{tabular}}
\\

\textit{Schaffers high cond}
& \begin{tabular}[c]{@{}c@{}}2.94e+01\\ $\pm$6.79e+00\end{tabular} 
& \begin{tabular}[c]{@{}c@{}}2.91e+01\\ $\pm$5.96e+00\end{tabular} 
& \begin{tabular}[c]{@{}c@{}}2.96e+01\\ $\pm$6.09e+00\end{tabular} 
& \begin{tabular}[c]{@{}c@{}}2.98e+01\\ $\pm$5.67e+00\end{tabular} 
& \textbf{\begin{tabular}[c]{@{}c@{}}2.68e+01\\ $\pm$5.99e+00\end{tabular}}
& \begin{tabular}[c]{@{}c@{}}2.37e+00\\ $\pm$1.97e+00\end{tabular} 
& \textbf{\begin{tabular}[c]{@{}c@{}}2.21e+00\\ $\pm$1.55e+00\end{tabular}}
& \begin{tabular}[c]{@{}c@{}}3.24e+00\\ $\pm$2.33e+00\end{tabular} 
& \begin{tabular}[c]{@{}c@{}}\underline{2.42e+00}\\ $\pm$2.54e+00\end{tabular} 
& \begin{tabular}[c]{@{}c@{}}3.09e+00\\ $\pm$2.18e+00\end{tabular} 
& \begin{tabular}[c]{@{}c@{}}1.21e+00\\ $\pm$6.64e-01\end{tabular} 
& \textbf{\begin{tabular}[c]{@{}c@{}}1.16e+00\\ $\pm$6.19e-01\end{tabular}}
& \begin{tabular}[c]{@{}c@{}}\underline{1.16e+00}\\ $\pm$6.23e-01\end{tabular}
& \begin{tabular}[c]{@{}c@{}}1.21e+00\\ $\pm$6.47e-01\end{tabular} 
& \begin{tabular}[c]{@{}c@{}}1.20e+00\\ $\pm$6.39e-01\end{tabular} 
\\

\textit{Schwefel}
& \begin{tabular}[c]{@{}c@{}}6.64e+03\\ $\pm$3.14e+03\end{tabular} 
& \begin{tabular}[c]{@{}c@{}}5.96e+03\\ $\pm$2.77e+03\end{tabular} 
& \begin{tabular}[c]{@{}c@{}}6.06e+03\\ $\pm$3.00e+03\end{tabular} 
& \begin{tabular}[c]{@{}c@{}}\underline{5.66e+03}\\ $\pm$2.52e+03\end{tabular} 
& \textbf{\begin{tabular}[c]{@{}c@{}}5.45e+03\\ $\pm$3.12e+03\end{tabular}}
& \begin{tabular}[c]{@{}c@{}}1.29e+00\\ $\pm$3.04e-01\end{tabular} 
& \begin{tabular}[c]{@{}c@{}}\underline{1.31e+00}\\ $\pm$2.90e-01\end{tabular} 
& \begin{tabular}[c]{@{}c@{}}1.42e+00\\ $\pm$2.54e-01\end{tabular} 
& \textbf{\begin{tabular}[c]{@{}c@{}}1.26e+00\\ $\pm$3.07e-01\end{tabular}}
& \begin{tabular}[c]{@{}c@{}}1.43e+00\\ $\pm$3.09e-01\end{tabular} 
& \begin{tabular}[c]{@{}c@{}}\underline{6.06e-01}\\ $\pm$2.59e-01\end{tabular}
& \begin{tabular}[c]{@{}c@{}}6.09e-01\\ $\pm$2.48e-01\end{tabular} 
& \begin{tabular}[c]{@{}c@{}}6.07e-01\\ $\pm$2.49e-01\end{tabular} 
& \begin{tabular}[c]{@{}c@{}}6.07e-01\\ $\pm$2.43e-01\end{tabular} 
& \textbf{\begin{tabular}[c]{@{}c@{}}6.06e-01\\ $\pm$2.48e-01\end{tabular}}
\\

\textit{Sharp Ridge}
& \begin{tabular}[c]{@{}c@{}}1.01e+03\\ $\pm$1.46e+02\end{tabular} 
& \begin{tabular}[c]{@{}c@{}}9.52e+02\\ $\pm$1.56e+02\end{tabular} 
& \begin{tabular}[c]{@{}c@{}}9.89e+02\\ $\pm$1.48e+02\end{tabular} 
& \begin{tabular}[c]{@{}c@{}}\underline{9.76e+02}\\ $\pm$1.46e+02\end{tabular} 
& \textbf{\begin{tabular}[c]{@{}c@{}}9.08e+02\\ $\pm$1.59e+02\end{tabular}}
& \textbf{\begin{tabular}[c]{@{}c@{}}1.10e+01\\ $\pm$1.17e+01\end{tabular} }
& \begin{tabular}[c]{@{}c@{}}1.92e+01\\ $\pm$3.43e+01\end{tabular} 
& \begin{tabular}[c]{@{}c@{}}1.62e+01\\ $\pm$2.97e+01\end{tabular} 
& \begin{tabular}[c]{@{}c@{}}\underline{1.13e+01}\\ $\pm$1.40e+01\end{tabular}
& \begin{tabular}[c]{@{}c@{}}1.55e+01\\ $\pm$2.78e+01\end{tabular} 
& \begin{tabular}[c]{@{}c@{}}2.36e+00\\ $\pm$2.03e+00\end{tabular} 
& \begin{tabular}[c]{@{}c@{}}\underline{2.27e+00}\\ $\pm$2.18e+00\end{tabular} 
& \begin{tabular}[c]{@{}c@{}}2.34e+00\\ $\pm$2.18e+00\end{tabular} 
& \textbf{\begin{tabular}[c]{@{}c@{}}2.19e+00\\ $\pm$2.20e+00\end{tabular}}
& \begin{tabular}[c]{@{}c@{}}2.32e+00\\ $\pm$2.05e+00\end{tabular} 
\\

\textit{Step Ellipsoidal}
& \begin{tabular}[c]{@{}c@{}}1.05e+02\\ $\pm$3.38e+01\end{tabular} 
& \begin{tabular}[c]{@{}c@{}}1.02e+02\\ $\pm$3.07e+01\end{tabular} 
& \begin{tabular}[c]{@{}c@{}}\underline{9.97e+01}\\ $\pm$3.06e+01\end{tabular} 
& \textbf{\begin{tabular}[c]{@{}c@{}}9.68e+01\\ $\pm$3.09e+01\end{tabular}}
& \begin{tabular}[c]{@{}c@{}}1.09e+02\\ $\pm$2.85e+01\end{tabular} 
& \begin{tabular}[c]{@{}c@{}}2.00e+00\\ $\pm$1.62e+00\end{tabular} 
& \begin{tabular}[c]{@{}c@{}}\underline{1.43e+00}\\ $\pm$9.46e-01\end{tabular} 
& \begin{tabular}[c]{@{}c@{}}2.71e+00\\ $\pm$2.10e+00\end{tabular} 
& \textbf{\begin{tabular}[c]{@{}c@{}}1.33e+00\\ $\pm$9.45e-01\end{tabular}}
& \begin{tabular}[c]{@{}c@{}}2.52e+00\\ $\pm$1.95e+00\end{tabular} 
& \begin{tabular}[c]{@{}c@{}}\underline{5.62e-01}\\ $\pm$4.98e-01\end{tabular} 
& \begin{tabular}[c]{@{}c@{}}6.16e-01\\ $\pm$5.66e-01\end{tabular} 
& \begin{tabular}[c]{@{}c@{}}5.62e-01\\ $\pm$4.99e-01\end{tabular} 
& \begin{tabular}[c]{@{}c@{}}6.02e-01\\ $\pm$5.52e-01\end{tabular} 
& \textbf{\begin{tabular}[c]{@{}c@{}}5.42e-01\\ $\pm$4.92e-01\end{tabular}}
\\ \hline
% ================= Win/Tie/Loss Statistics =================
\hline
\textit{Avg. Rank}
% DEDQN: Handcrafted, EoH, ReEvo, Eureka, READY
& 4.06 & 2.31 & 3.44 & 3.19 & \textbf{2.00}
% RLEPSO: Handcrafted, EoH, ReEvo, Eureka, READY
& 2.75 & 3.44 & 3.06 & 3.38 & \textbf{2.38}
% RLDAS: Handcrafted, EoH, ReEvo, Eureka, READY
& 3.31 & 2.93 & 3.56 & 3.13 & \textbf{2.06}
\\ \hline
% \hline
% \textbf{} $^\dagger$
% % DEDQN (READY vs Baseline)
% & 14/0/2 & 10/0/6 & 12/0/4 & 12/0/4 & - 
% % RLEPSO (READY vs Baseline)
% & 10/0/6 & 10/0/6 & 12/0/4 & 10/0/6 & - 
% % RLDAS (READY vs Baseline)
% & 12/1/2 & 9/2/5 & 12/1/3 & 10/0/6 & - 
% \\ \hline
\end{tabular}%
}
\vspace{-3mm}
\end{table*}
% =================================================================
% SECTION 4: EXPERIMENTS
% =================================================================
\section{Experiments}
\label{sec:experiments}

\subsection{Experimental Setup}

\paragraph{Multitask MetaBBO Environment.}
To evaluate the multitasking performance of READY, we construct a multitask environment comprising three distinct MetaBBO:
\begin{itemize}
    \item \textbf{RLDAS~\cite{guo2024deep}} (Algorithm Selection): A discrete control task where a PPO agent dynamically switches different Differential Evolution algorithms during optimization.
    \item \textbf{RLEPSO~\cite{yin2021rlepso}} (Parameter Control): A continuous control task employing PPO for the adaptive adjustment of evolutionary factors in Particle Swarm Optimization.
    \item \textbf{DEDQN~\cite{tan2021differential}} (Operator Selection): A discrete control task relying on DQN to select mutation operators from a predefined candidate pool for Differential Evolution.
\end{itemize}

\paragraph{Benchmarks and Baselines.}
We employ the BBOB test suite~\cite{ma2025metabox} with a strict train-test split: 8 function are used for training, while 16 unseen problems are reserved for zero-shot testing. We benchmark READY against handcrafted rewards and two categories of automated baselines: (1) the dedicated reward discovery method \textbf{Eureka}~\citep{ma2023eureka}; and (2) general LLM-based algorithm design frameworks \textbf{EoH}~\citep{liu2024evolution} and \textbf{ReEvo}~\citep{ye2024reevo}, adapted for reward discovery. Notably, while READY discovers rewards for these three heterogeneous environments simultaneously within a single unified run, all baseline methods are trained independently for each specific MetaBBO.

\paragraph{Implementation Details.}
 % We use DeepSeek-V3.2 as the LLM backbone, setting niche size $N=5$ and generations $G=7$. To balance accuracy and speed under a strict 0.5-hour training budget, we utilize Ray for distributed parallel acceleration, evaluating each candidate reward across 3 independent runs concurrently.
We utilize DeepSeek-V3.2~\cite{liu2024deepseek} as the LLM backbone, setting the niche size as $N=5$ and the maximum evolutionary generations as $G_\text{max}=7$. To balance accuracy and speed, we impose a 0.5-hour wall-clock limit on the training of MetaBBO. We utilize the Ray framework to accelerate the fitness assessment via parallel evaluation. During the evolutionary search, each reward is tested over $\Gamma=3$ independent runs to mitigate stochasticity.
 
\begin{figure*}
    \centering
    \includegraphics[width=1\linewidth]{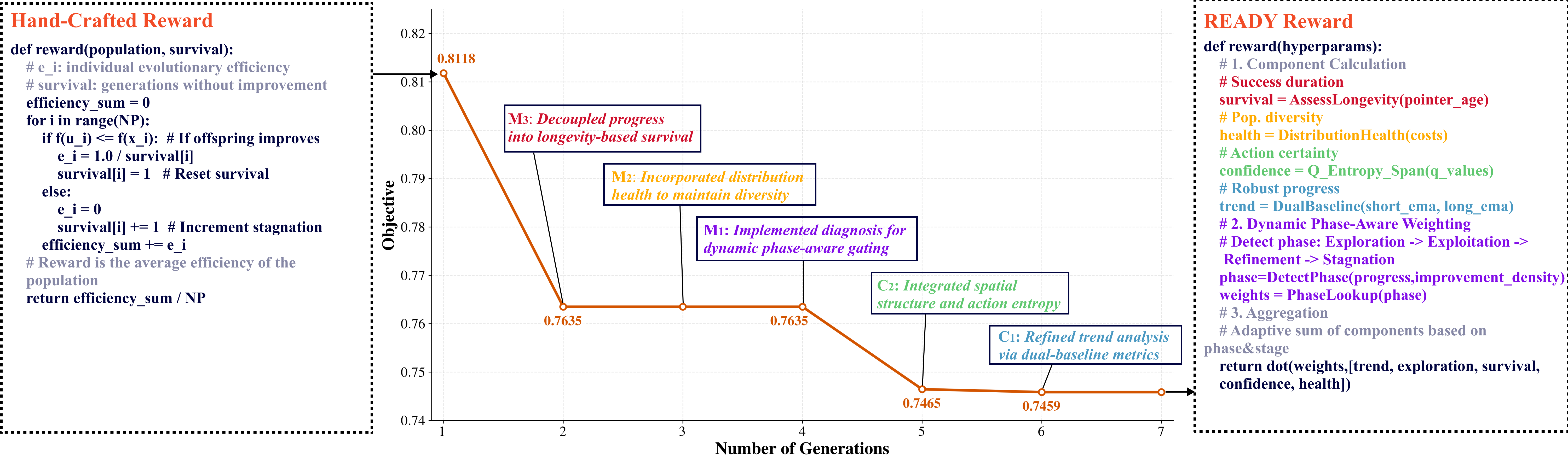}
    \caption{Evolutionary trajectory of READY's best-so-far performance when discover reward for DEDQN. The plateau reflects a phase of strategic accumulation, serving as a necessary precursor for the significant exploitation breakthrough.}
    \vspace{-3mm}
    \label{fig:evolution_trajectory}
\end{figure*}
\subsection{Main Experimental Results}
\label{subsec:main_results}
Following the reward discovery, we remove the training time constraints to fully train the optimal rewards from READY and the baselines. Each reward is then evaluated across 51 independent runs on the same test suite to ensure consistency.
Table~\ref{tab:main_res} reports the mean minimum cost achieved by each method across three MetaBBO frameworks.
\paragraph{Overall Optimization Capability.}
As summarized in Table~\ref{tab:main_res}, READY demonstrates superior generalization capabilities. Despite being trained within a unified framework, READY outperforms the Handcrafted rewards in 36 out of 48 test scenarios (covering 16 functions across 3 tasks). Furthermore, when compared against the full suite of state-of-the-art automated baselines, READY achieves the lowest mean cost in 24 cases. This result is particularly significant given that baseline methods were trained independently for each specific MetaBBO environment, whereas READY utilized a single set of discovered rewards to generalize across all diverse decision spaces simultaneously.
\paragraph{Versatility Across Heterogeneous Agents.}
READY exhibits remarkable paradigm-agnostic adaptability. As shown in the \textit{Attractive Sector} and \textit{Different Powers} functions, READY achieves top-tier performance across all three distinct control topologies. For instance, on \textit{Attractive Sector}, READY secures the best performance on DEDQN ($2.35 \times 10^3$) and RLDAS ($1.28 \times 10^{-1}$), while maintaining competitive results on RLEPSO. In contrast, Handcrafted rewards often fail to generalize to the continuous dynamics of RLEPSO (e.g., significantly underperforming on \textit{Ellipsoidal High Cond}), highlighting READY's ability to evolve universal signaling mechanisms that transcend specific agent architectures.

\paragraph{Robustness on Complex Landscapes.}
We observe that READY effectively balances the exploration-exploitation trade-off, specifically tailoring its strategy to the optimization context. On highly deceptive, multimodal functions such as \textit{Buche Rastrigin} and \textit{Schwefel}, READY demonstrates superior global search capabilities, particularly within the DEDQN task where it significantly outperforms baselines (e.g., achieving a minimum cost of $3.18 \times 10^2$ on \textit{Buche Rastrigin}). This exploration capability is complemented by distinct exploitation strength on ill-conditioned, unimodal landscapes. A standout example is \textit{Ellipsoidal High Conditional}, where READY achieves best performance across all three MetaBBO. Notably, on RLEPSO, READY reduces the cost by approximately 38\% compared to Eureka ($5.71 \times 10^2$ vs. $9.21 \times 10^2$). These results confirm that READY does not rely on a single static heuristic but dynamically evolves specialized reward logics adapted to the distinct topological requirements of each landscape.

\paragraph{Stability and Reproducibility.}
READY also enhances training stability, frequently reporting lower standard deviations than baselines. This reduced variance indicates robustness to initialization noise and more reproducible RL trajectories.

\subsection{Trajectory Analysis: From Accumulation to Breakthrough}
\label{subsec:evolutionary_trajectory}

Figure \ref{fig:evolution_trajectory} visualizes the evolutionary trajectory on the DEDQN task, revealing a distinct Exploration-Accumulation-Exploitation pattern that diverges from linear convergence. As detailed below, specific operators drive three critical phases:
\begin{itemize}
    \item \textbf{Rapid Adaptation (Gen 1-2):} The $M_3$ mutation operator drives an early performance spike by shifting focus from sparse individual signals to population-averaged efficiency, establishing a strong initial baseline.
    \item \textbf{Latent Accumulation (Gen 2-4):} A strategic plateau emerges, representing a phase of \textit{structural refinement}. $M_2$ and $M_1$ mutation operators debug landscape-specific failures and refine trade-off mechanisms. While global loss remains stable, the reward function accumulates complexity and robustness.
    \item \textbf{Breakthrough (Gen 5+):} This latent potential is catalyzed by the crossover operators, which integrates spatial-entropic feedback to trigger a breakthrough, successfully escaping local optima.
\end{itemize}
The final evolved reward for DEDQN synthesizes a phase-adaptive mechanism that dynamically reweights five strategic components, ranging from trend improvement to exploration according to real-time search states, ensuring consistent convergence across diverse landscapes.

\subsection{Computational Efficiency}
\label{subsec:efficiency}
We evaluate the efficiency of READY from two perspectives: the evolutionary search time and the inference latency.
\paragraph{Evolutionary Search Efficiency.}
Benefiting from the niche-based multi-task architecture, READY significantly reduces the wall-clock time required for reward discovery. While baseline methods must be executed independently for each MetaBBO task, READY concurrently evolves specialized rewards for all tasks in a single unified run. READY completes the entire discovery process across three tasks in only 7 hours. In contrast, sequential execution of Eureka requires approximately 14 hours, while EoH and ReEvo take over 14.6 hours and 30.6 hours respectively to achieve comparable coverage. This represents a 2$\times$ to 4$\times$ speedup in search efficiency, validating that cross-niche knowledge sharing and parallel evaluation effectively accelerate evolutionary convergence.
\paragraph{Inference Latency.}
To ensure practical applicability, we profiled the runtime latency of the discovered rewards. Empirical measurements confirm that despite their internal symbolic complexity, the rewards operate strictly within the microsecond regime (e.g., as low as $3.47 \mu$s per call for DEDQN). Although this introduces a slight overhead compared to handcrafted baselines, the cost is negligible as objective function evaluations typically dominate the BBO runtime which often requiring milliseconds to minutes. 
\begin{table}[h]
\centering
\caption{Zero-shot transfer performance. Objective values of the READY-evolved DEDQN reward when applied to RLDEAFL.}
\resizebox{0.8\columnwidth}{!}{%
\begin{tabular}{lccc}
\toprule
Problem & RLDEAFL & READY & Reduction (\%) \\
\midrule
\textit{Discus} & 6.48e+00 & \textbf{4.05e-04} & 99.99 \\
\textit{Ellipsoidal high cond} & 3.50e+02 & \textbf{2.57e-01} & 99.93 \\
\textit{Attractive Sector} & 1.20e-01 & \textbf{4.99e-04} & 99.59 \\
\textit{Different Powers} & 4.26e-04 & \textbf{1.77e-05} & 95.84 \\
\textit{Sharp Ridge} & 1.95e+00 & \textbf{8.35e-02} & 95.71 \\
\textit{Bent Cigar} & 5.89e+00 & \textbf{6.17e-01} & 89.54 \\
\textit{Rosenbrock original} & 4.13e+00 & \textbf{1.34e+00} & 67.61 \\
\textit{Rosenbrock rotated} & 5.69e+00 & \textbf{1.88e+00} & 66.95 \\
\textit{Schaffers high cond} & 2.81e-01 & \textbf{9.90e-02} & 64.79 \\
\textit{Step Ellipsoidal} & 1.23e-01 & \textbf{7.26e-02} & 41.00 \\
\textit{Katsuura} & 1.32e+00 & \textbf{8.46e-01} & 35.87 \\
\textit{Lunacek bi Rastrigin} & 3.64e+01 & \textbf{2.53e+01} & 30.37 \\
\textit{Composite Grie rosen} & 2.47e+00 & \textbf{2.35e+00} & 4.80 \\
\textit{Buche Rastrigin} & \textbf{1.77e+01} & 1.98e+01 & -11.89 \\
\textit{Gallagher 21Peaks} & \textbf{1.52e+00} & 2.47e+00 & -62.59 \\
\textit{Schwefel} & \textbf{2.77e-01} & 6.55e-01 & -136.80 \\
\bottomrule
\end{tabular}
}
\label{tab:transfer_comparison}
\end{table}
\subsection{Generalizability via Zero-Shot Transfer}
\label{subsec:transfer}

To investigate whether READY captures intrinsic optimization principles rather than overfitting to specific architectures, we evaluated the zero-shot transfer of reward logic across distinct MetaBBO  frameworks via LLM-based translation without fine-tuning. As detailed in Table~\ref{tab:transfer_comparison}, transferring logic from DEDQN to RLDEAFL~\cite{guo2025reinforcement} yields performance improvements on 13 out of 16 functions, achieving up to 99.99\% cost reduction despite the significant structural gap between source and target tasks. This generalizability is further corroborated by a secondary experiment transferring from RLEPSO to GLEET~\cite{ma2024auto}, which improved performance on 11 out of 16 instances. The results validate that READY identifies universal, algorithm-agnostic heuristics in discovering rewards.

\begin{figure}[t]
    \centering
    \includegraphics[width=1\linewidth]{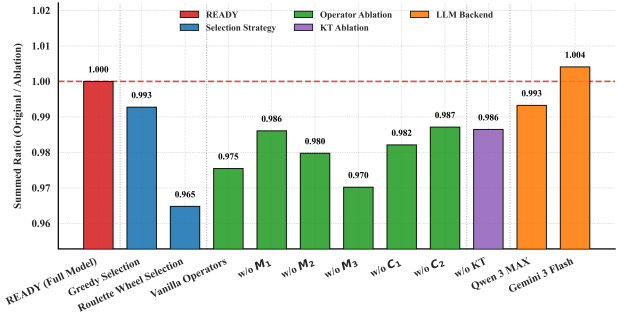}
    \caption{Ablation and sensitivity analysis. Y-axis shows Summed Normalized Efficiency, with the red dashed line ($y=1.0$) marking the full model baseline. }%\textbf{Left:} Consistent drops in all variants validate the necessity of each semantic operator. \textbf{Right:} Results across backbones confirm robustness and positive scaling with reasoning capability.}
    \vspace{-3mm}
    \label{ablation_fig}
\end{figure}
\subsection{Ablation Study and Sensitivity Analysis}
\label{subsec:ablation}
To validate component contributions, we conducted ablation studies quantified by the Summed Normalized Efficiency (SNE). The SNE of baseline $B$ is defined as: 
% \begin{equation}
%     SNE_B = \frac{1}{K}\sum_{k=1}^K \frac{\mathcal{F}(\text{READY} | \mathcal{T}_k)}{\mathcal{F}(B | \mathcal{T}_k)}
% \end{equation}
$SNE_B = \frac{1}{K}\sum_{k=1}^K \frac{\mathcal{F}(\text{READY} | \mathcal{T}_k)}{\mathcal{F}(B | \mathcal{T}_k)}$. 
A score of 1.0 signifies parity with the full READY framework, while lower scores indicate performance degradation.

\paragraph{Component Efficacy via Iso-Budget Ablation.}
We substitute target  operators with simple LLM mutations to maintain a constant evaluation budget. Figure \ref{ablation_fig} reveals that replacing any specialized operator causes a significant drop in SNE. This confirms that READY's efficacy stems from its specific architectural logic rather than the mere quantity of inferences. Disabling the Knowledge Transfer module degrades performance, confirming that sharing heuristics across niches is essential for effective reward discovery.

\paragraph{Scalability with Reasoning Capability.}
We evaluated robustness across diverse backbones: DeepSeek-V3.2~\cite{liu2024deepseek}, Qwen-3-Max~\cite{yang2025qwen3}, and Gemini-3-Flash~\cite{team2023gemini}. READY delivers consistent high-quality solutions across all models. Crucially, the superior performance achieved by Gemini suggests a positive scaling law: the framework is not architecturally bottlenecked but scales effectively with the backend's reasoning capability, indicating strong potential to leverage future advancements in foundation models.

\section{Conclusion}
\label{sec:conclusion}

We proposed READY, the first LLM-driven multitask framework for automating reward discovery in MetaBBO. By synergizing a niche-based architecture with specialized LLM operators, READY overcomes the limitations of manual reward engineering in effectiveness and efficiency. Empirical results confirm that READY generates high-quality, interpretable rewards that outperform baselines and exhibit robust zero-shot generalizability across diverse tasks. This framework paves the way for fully autonomous optimization pipelines by capturing universal search heuristics.

\section*{Impact Statement}
This paper presents work whose goal is to advance the field of MetaBBO. We propose a reward design framework to automate the reward design process in establishing MetaBBO systems. This is a significant step forward and hance may produces profound impacts on the future development of MetaBBO researches, especially in two aspects: 1) the next generation self-organized learning agent such as automated MetaBBO framework design; 2) boosting MetaBBO's performance on realworld applications.

\bibliography{ref}
\bibliographystyle{icml2026}

%%%%%%%%%%%%%%%%%%%%%%%%%%%%%%%%%%%%%%%%%%%%%%%%%%%%%%%%%%%%%%%%%%%%%%%%%%%%%%%
%%%%%%%%%%%%%%%%%%%%%%%%%%%%%%%%%%%%%%%%%%%%%%%%%%%%%%%%%%%%%%%%%%%%%%%%%%%%%%%
% APPENDIX
%%%%%%%%%%%%%%%%%%%%%%%%%%%%%%%%%%%%%%%%%%%%%%%%%%%%%%%%%%%%%%%%%%%%%%%%%%%%%%%
%%%%%%%%%%%%%%%%%%%%%%%%%%%%%%%%%%%%%%%%%%%%%%%%%%%%%%%%%%%%%%%%%%%%%%%%%%%%%%%
\newpage
\appendix
\onecolumn

\section{Detailed Prompt}
\label{app:prompts}

We provide the full prompt templates used for the semantic evolutionary operators.

% ==========================================

% INIT PROMPT

% ==========================================
\subsection{Initialization Prompt}
\label{initprompt}
\begin{dialogbox}
\textbf{\# Role}
You are a reward engineer trying to write reward functions to solve Metablackbox Optimization tasks as effective as possible.

\textbf{\# Aim}
Your goal is to write a reward function for the environment that will help the agent learn the task described in text, and perform well on the problems.

Your reward should use useful variables from the environment as inputs.

\textbf{\# Info}

You are working on the task:\{Task\_description\}

Experts have proposed \{MetaBBO\_rewards\_nums\} Metablackbox Optimization rewards to solve this problem. The ideas for these Metablackbox Optimization rewards are as follows:

Please create a new reward that differs from the existing reward by at least \{difference\_rate\} \%

The new reward should perform better than \textbf{the existing rewards} on the given task. (If not, try your best to make it better than \textbf{the original reward})

\end{dialogbox}
% ==========================================
% FRE - Phase 1
% ==========================================

\subsection{$M_1$ Local-Reflection Mutation Prompt}
\label{app:m1prompt}
\subsubsection{Reflection Phase}
\begin{dialogbox}
\textbf{\# Role}

You are a MetaBBO algorithm reward design expert specializing in \textbf{failure analysis}.

\textbf{\# Info}

An intelligent agent is currently executing the following design task:\{Task\_description\}

The agent has designed a reward function that performs well on some problems but \textbf{fails significantly on specific problems}.

\textbf{\# Current Reward Info}

\textbf{Original Thought:}\{Thought\}

\textbf{Original Code:}\{Code\}

\textbf{\# Failure Analysis Report}

We have identified a set of problems where this reward function performs poorly. The characteristics of these problems are described below:

\textbf{Characteristics of Poorly Performing Problems:}\{Bad\_Case\_Characteristics\}

\textbf{Performance Gap:}

On these problems, The reward's performances are:\{Bad\_Case\_Performance\}

\textbf{\# Your Task}

Based on your understanding and knowledge, please provide suggestions for the current reward to guide the agent in improving it.

The suggestion should include:
\begin{enumerate}
    \item \textbf{Diagnose}: Analyze WHY the current reward logic (based on the Original \texttt{Thought} and \texttt{Code}) is insufficient for the problems described in the \texttt{Failure Analysis Report}.
    \item \textbf{Propose Solution}: Briefly describe potential the modification strategy.
    \begin{itemize}
        \item \textbf{CONTENT}: Provide the conceptual direction. Describe \textit{what} to change, not \textit{how} to code it. \textbf{Do not include Python code or complex derivations.}
    \end{itemize}
\end{enumerate}
\end{dialogbox}
% ==========================================
% FRE - Phase 2
% ==========================================
\subsubsection{Mutation Phase}
\begin{dialogbox}
\textbf{\# Role}

You are a Metablackbox Optimization reward design expert. 

Remember: \textbf{Lower Fitness is Better}. A decrease in the function value indicates improvement.

\textbf{\# Info}

An intelligent agent is currently executing the following design task:\{Task\_description\}

The agent has designed a reward with the following code:\{Code\}

\textbf{\# Aim}

An expert has provided some suggestions for this Metablackbox Optimization reward. You can decide how to incorporate the advice, and create a new reward that differs from the given one but motivated by it.

The advice is: \{Reflection\}

\end{dialogbox}
% ==========================================

% HGE

% ==========================================
\subsection{$M_2$ History-Reflection Prompt}
\label{app:m2prompt}
\begin{dialogbox}

\textbf{\# Role}

You are an expert in evolutionary MetaBBO and reward function design. Remember: \textbf{Lower Fitness is Better}. A decrease in the function value indicates improvement.

\textbf{\#\# Task}

You are currently optimizing a reward function for the following task:

\{Task\}

\textbf{\#\#\# Evolutionary Trajectory Analysis}

I will provide you with the \textbf{evolutionary history} of the current individual. This history shows how the reward function has improved step-by-step. Your goal is to \textbf{identify the optimization trend} and \textbf{extrapolate the next logical improvement}.

\textbf{\#\#\# History Trace}

\{History\_Trace\}

\textbf{\#\#\# Current Individual}

\textbf{Fitness:} \{Current\_Fitness\}

\textbf{Fitness Detailed:}\{Fitness\_Detailed\}

\textbf{Thought:}\{Current\_Thought\}

\textbf{Code:}\{Current\_Code\}

\textbf{\#\#\# Your Task}
\begin{enumerate}
    \item Analyze the Trajectory: Briefly explain what changed from the past versions to the current version and why it led to performance improvements. What is the underlying direction? 
    \item Extrapolate: Based on this trend, propose the next step. Don't just randomly mutate; follow the momentum of the history. If the previous steps successfully refined a specific component, continue refining it or address the side effects caused by it.
    \item Generate Code: Output the new reward function code.
\end{enumerate}
\end{dialogbox}
% ==========================================
% SE
% ==========================================
\subsection{$M_3$ Global-Reflection prompt}
\label{app:m3prompt}
\subsubsection{Reflection Phase}
\begin{dialogbox}
\textbf{\# Role}

You are the person responsible for recording the progress of the experts’ research.

You are working on the following task:\{Task\_Description\}

\textbf{\#\# Info}

Experts have now explored an additional \{MetaBBO\_reward\_nums\} Metablackbox Optimization rewards, ranging from No. 1 to No. \{MetaBBO\_reward\_nums\}:

\textbf{\# Aim}

Please review the previous experiences and the current Metablackbox Optimization rewards, analyzing which techniques within the Metablackbox Optimization rewards are effective in solving this problem and which are not. Finally, summarize both the effective and ineffective techniques, and update the previous summary accordingly. Please ensure that the summary you provide is written as:

\begin{verbatim}
```summary
{summary}
```
\end{verbatim}
\end{dialogbox}

\subsubsection{Mutation Phase}

\begin{dialogbox}
\textbf{\# Role}

You are an Metablackbox Optimization reward design expert, currently collaborating with other experts on the following task:\{Task\}

\textbf{\# Info}

Currently, \{reinforcement\_learning\_reward\_nums\} Metablackbox Optimization rewards have been explored for this problem, with their effectiveness decreasing from No. 1 to No. \{reinforcement\_learning\_reward\_nums\}. The concepts for these methods are as follows.

The summary of the archive is: \{summary\}

Please analyze the summary and then modify the following reward to create a more promising reward. The thoughts and code for the reward to be modified are as follows:

\{individual.thought\}

\{individual.reward\_code\}
\end{dialogbox}

% ==========================================
% ESE
% ==========================================
\subsection{$C_1$ Exploitative prompt}
\label{app:c1prompt}
\begin{dialogbox}
\textbf{\# Role}

You are an Metablackbox Optimization reward design expert.
Remember: \textbf{Lower Fitness is Better}. A decrease in the function value indicates improvement.

\textbf{\# Info}

Experts are divided into several groups, with each group responsible for the development of a specific Metablackbox Optimization reward cluster. Each cluster incorporates different techniques while maintaining its own framework to explore diverse Metablackbox Optimization rewards.

On your Metablackbox Optimization reward cluster, after several iterations, the current Metablackbox Optimization reward (idea and the corresponding code) is:\{chosen\_MetaBBO\_reward\_thought\},\{chosen\_MetaBBO\_reward\_code\}

During the iterations in your cluster, a better-performing Metablackbox Optimization reward appeared, and its idea and code are as follows:\{cluster\_best\_MetaBBO\_reward\_thought\},\{cluster\_best\_MetaBBO\_reward\_code\}

In addition, among all the Metablackbox Optimization rewards tested (including those from other clusters), the best-performing Metablackbox Optimization reward’s idea and code are as follows:\{global\_best\_MetaBBO\_reward\_thought\},\{global\_best\_MetaBBO\_reward\_code\}

\textbf{\# Aim}

Using the above information and adhering to the core framework of the current Metablackbox Optimization reward, please suggest potential improvements to enhance its performance in solving this problem.

There are some info that would restric your design of the reward:

The lab is now currently collaborating with other experts on the following task:\{Task\_description\}

\end{dialogbox}

% ==========================================
% ECC
% ==========================================
\subsection{$C_2$ Exploratory prompt}
\label{app:c2prompt}
\begin{dialogbox}

\textbf{\# Role}

You are an Metablackbox Optimization reward design expert, currently collaborating with other experts on the following task:\{Task\_description\}

\textbf{\# Info}

Experts have designed \{MetaBBO\_reward\_nums\} Metablackbox Optimization rewards with their corresponding codes.

The No. 1 Reward and the corresponding code are:\{individual1.thought\},\{individual1.reward\_code\}

The No. 2 Reward and the corresponding code are:\{individual2.thought\},\{individual2.reward\_code\}

Please take Reward No. 1 as the main framework and try to incorporate the characteristics of the other rewards into it to create a better reward.

\end{dialogbox}

\subsection{Knowledge Transfer Prompt}
\label{app:ktprompt}
\begin{dialogbox}
\textbf{\# Role}

You are an advanced \textbf{Optimization Meta-Learner}. Your goal is to orchestrate knowledge transfer between different evolutionary optimization tasks to optimize their performance. Remember: \textbf{Lower Fitness is Better}. A decrease in the function value indicates improvement.

\textbf{\# Aim}

Analyze the provided tasks and historical data to determine optimal \textbf{Task Pairs} for knowledge transfer.

\textbf{Core Mechanism:} The system will \textbf{automatically} extract the \textbf{Best Individual} from the chosen Source Task and adapt it to replace the \textbf{Worst Individual} in the Target Task.

\textbf{Your Focus:} You do NOT need to select specific individuals. Your sole responsibility is to identify which \textbf{Source Task} (Algorithm/Problem) possesses knowledge that is valuable for a specific \textbf{Target Task}.

\textbf{\# Data Overview}

\textbf{\#\# 1. Historical Knowledge}

This data reflects the outcome of your past decisions. Use it to learn what works.\{kt\_historical\_info\}

\textbf{\#\# 2. Active Tasks \& States}

Below are the descriptions and the current internal thoughts/states of the top individuals in each task.
\{tasks\_descriptions\}

\textbf{\# Instruction}

Determine the \{N\_direction\} most valuable transfer operations (Source Task -> Target Task).

\textbf{CRITICAL RULES:}
\begin{enumerate}
    \item \textbf{Focus on Task Compatibility:} Do NOT base your decision on the specific traits of an individual. Instead, base it on the \textbf{Algorithm/Task characteristics}.
    \item \textbf{Fixed Mechanism:} Remember that the "Best Individual" is automatically selected.
    \item \textbf{Strategy:} For each operation, generate a \textbf{Strategy} that describes \textit{how} to transform the logic/code from Source to Target to ensure compatibility.
\end{enumerate}

\textbf{\# Output Format (JSON)}

Please output strictly in the following JSON format:

\begin{lstlisting}
[
    {{
        "source_task_Metabbo_algorithm": "{{source_algorithm_name}}",
        "target_task_Metabbo_algorithm": "{{target_algorithm_name}}",
        "rationale": "Explain **WHY** you chose this pair",
        "transfer_strategy_guidance": "Specific instruction on **HOW** to map the individual/code."
    }},
    ...
]
\end{lstlisting}
\end{dialogbox}

\section{Specific MetaBBO MetaData}
\subsection{DEDQN MetaData}
\begin{dialogbox}
\textbf{DEDQN (Differential Evolution with Deep Q-Network)}

The \textbf{DEDQN} algorithm primarily focuses on the \textbf{adaptive selection of mutation strategies in Differential Evolution (DE)}. It aims to solve the problem of selecting the most appropriate mutation strategy. The algorithm uses a Deep Q-Network (DQN) to manage a mixed mutation strategy pool during the evolution process.

\textbf{Algorithm Framework}

DEDQN operates in two distinct phases: an offline training phase and an online prediction phase.
\begin{itemize}
    \item \textbf{Offline Training Phase}:
    \begin{itemize}
        \item In this stage, the DQN is \textbf{trained offline}.
        \item It learns by collecting data from multiple DE runs on various training functions, establishing a relationship between the "fitness landscape" features (state) and the "benefit (reward)" of applying each mutation strategy (action).
        \item The DQN acts as an agent, interacting with an environment composed of the DEDQN algorithm, the fitness function, and a module for calculating the fitness landscape.
    \end{itemize}
    \item \textbf{Online Prediction Phase}:
    \begin{itemize}
        \item After training, the DQN's neural network weights are fixed.
        \item When the DEDQN algorithm is applied to solve a new (test) problem, it uses this trained DQN.
        \item \textbf{At each generation}, the algorithm calculates the current fitness landscape features and feeds them to the DQN, which then \textbf{predicts} and selects the most suitable mutation strategy to be used.
    \end{itemize}
\end{itemize}

\textbf{Key Features \& Mechanisms}
\begin{itemize}
    \item \textbf{Mixed Mutation Strategy Pool (Action Space)}: The DEDQN's action space consists of three DE mutation strategies, each chosen for its different search characteristics:
    \begin{itemize}
        \item \textbf{"DE/rand/1"} (for local search)
        \item \textbf{"DE/current to rand/1"} (for global search)
        \item \textbf{"DE/best/2"} (for rapid convergence)
    \end{itemize}
    \item \textbf{Population Evolutionary Efficiency (Reward)}: The reward signal used to train the DQN is defined as the "\textbf{population evolutionary efficiency}". This metric measures the evolutionary ability of the population based on individual "survival" from one generation to the next.
    \item \textbf{Parameter Adaptation}: In addition to the DQN-based mutation strategy selection, DEDQN also employs a "\textbf{historical memory parameter adaptation mechanism}" (similar to that in SHADE) to adapt the $F$ and $Cr$ control parameters.
\end{itemize}

\textbf{The reward\_hyperparameters could be used in DEDQN}

\begin{lstlisting}[language=Python, basicstyle=\ttfamily\scriptsize, breaklines=true]
reward_hyperparameters = {
    # From Optimizer (DEDQN)
    "survival": np.ndarray,          # survival counts per individual (length NP)
    "pointer": int,                  # index of the individual being updated
    "population": np.ndarray,        # population array, shape [NP, dim]
    "costs": np.ndarray,             # population costs, shape [NP,]
    "parent_cost": float,            # cost of the pointer individual before generating trial
    "trial_cost": float,             # cost of the trial (after mutation/crossover)
    "gbest_cost": float,             # current global best cost
    "median_cost": float,            # median of population costs
    "mean_cost": float,              # mean of population costs
    "std_cost": float,               # std of population costs
    "diversity": float,              # mean std across dimensions (approx diversity)
    "FEs": int,                      # current number of function evaluations
    "MaxFEs": int,                   # maximum allowed function evaluations
    "progress": float,               # normalized progress = FEs / MaxFEs in [0,1]
    "action": int,                   # chosen mutation strategy id (e.g. 0,1,2)
    "generation": int,               # current generation index (approx FEs // NP)
    "accepted": int,                 # whether trial replaced parent (1) or not (0)
    "delta_cost": float,             # parent_cost - trial_cost (>0 means improvement)
    "gbest_improve": float,          # previous gbest_cost - current gbest_cost
    "pointer_age": float,            # survival age of the current pointer individual

    # From Agent (DQN) via action payload (agent_context)
    "q_values": np.ndarray,          # Q(s, .), shape [n_env, n_act]
    "greedy_action": np.ndarray,     # argmax_a Q(s,a), shape [n_env,]
    "q_span": np.ndarray,            # per-env max(Q) - min(Q)
    "q_entropy": np.ndarray,         # per-env entropy of softmax(Q / tau)
    "recent_reward_mean": np.ndarray,# moving mean of last-K rewards per env
    "recent_reward_max": np.ndarray, # moving max of last-K rewards per env
    "training_step": int,            # RL training step counter
}
\end{lstlisting}
\end{dialogbox}

\subsection{RLDAS MetaData}
\label{rldasmetadata}
\begin{dialogbox}
\textbf{RL-DAS (Deep Reinforcement Learning-based Dynamic Algorithm Selection)}

The \textbf{RL-DAS} framework primarily focuses on \textbf{Dynamic Algorithm Selection (DAS) in Black-Box Optimization (BBO)}. It aims to address the limitation that a single algorithm's effectiveness varies across different problems. It does this by leveraging the complementary strengths of a group of algorithms and dynamically scheduling them throughout the optimization process for a specific problem.

\textbf{Algorithm Framework}

RL-DAS employs an architecture centered around a deep reinforcement learning agent that interacts with an optimization environment (which includes the problem, the population, and an algorithm pool).
\begin{itemize}
    \item \textbf{RL Agent}: This is the core decision-maker, trained using a policy gradient method, specifically \textbf{Proximal Policy Optimization (PPO)}. The entire process is modeled as a \textbf{Markov Decision Process (MDP)}.
    \item \textbf{Algorithm Pool}: This contains a finite set of $L$ candidate algorithms. In the paper's proof-of-principle study, this pool consists of three advanced Differential Evolution (DE) algorithms: JDE21, MadDE, and NL-SHADE-RSP.
    \item \textbf{Optimization Process}: The optimization is divided into time intervals. At each interval (or decision step) $t$:
    \begin{enumerate}
        \item The framework extracts a \textbf{State Feature} from the current problem and population.
        \item The RL agent receives this state and \textbf{Selects} an action $a_t$, which corresponds to choosing an algorithm from the pool.
        \item The chosen algorithm is executed for a period, updating the population.
        \item The framework observes a reward and transitions to the next state, repeating the process.
    \end{enumerate}
\end{itemize}

\textbf{Key Features \& Mechanisms}
\begin{itemize}
    \item \textbf{Dynamic Algorithm Scheduling}: Unlike traditional (static) Algorithm Selection (AS) which picks one algorithm for the entire run, RL-DAS dynamically switches between algorithms during the optimization process, allowing it to achieve a comprehensively better performance than the best single algorithm in the pool.
    \item \textbf{Algorithm Context Restoration}: A critical mechanism that enables \textbf{smooth switching} between algorithms. It uses a "context memory" Gamma to save and restore the algorithm-specific internal states (like adaptive parameters, statistical measures, or archives). This allows an algorithm to be \textbf{warm-started} from where it left off, rather than reinitializing.
    \item \textbf{Generality and Generalization}: The framework is designed to be simple and generic, offering potential improvements for a broad spectrum of evolutionary algorithms (EC), not just DE. Experiments showed it has favorable generalization ability, achieving strong performance in \textbf{zero-shot} scenarios on unseen problem classes.
\end{itemize}

\textbf{The reward\_hyperparameters could be used in RLDAS}

\begin{lstlisting}[language=Python, basicstyle=\ttfamily\scriptsize, breaklines=true]
reward_hyperparameters = {
    # Information from Optimizer 
    "last_cost": float,              # Global best cost before this step (population.gbest at step start)
    "current_gbest": float,          # Global best cost after this step (population.gbest at step end)
    "cost_scale_factor": float,      # Scale factor for costs (set to initial best cost at init)
    "FEs": int,                      # Current function evaluations used
    "MaxFEs": int,                   # Maximum function evaluations allowed
    "action": int,                   # Selected optimizer index in the pool (e.g., 0: NL_SHADE_RSP, 1: MadDE, 2: JDE21)
    "problem": object,               # Problem instance (access lb/ub/optimum/eval/func if needed)

    # Population handle (object) -- access fields as needed:
    #   population.group:        np.ndarray, shape [NP, dim]
    #   population.cost:         np.ndarray, shape [NP,]
    #   population.gbest:        float
    #   population.gbest_solution: np.ndarray, shape [dim,]
    #   population.archive:      np.ndarray, shape [<= NA, dim]
    #   population.NP / NA / dim: ints
    "population": object,

    # Information from Agent (trainer side)
    "agent_state": np.ndarray,       # State observed by the agent (batch/env-dependent)
    "policy_entropy": np.ndarray,    # Policy entropy (encourages exploration), shape depends on agent
    "value_estimation": np.ndarray,  # Critic value estimation
    "log_probability": np.ndarray,   # Log prob of taken action
    "gamma": float,                  # Agent's discount factor
    "learning_rate": float,          # Agent's current learning rate
}
\end{lstlisting}
\end{dialogbox}

\subsection{RLEPSO MetaData}
\label{rlepsometadata}
\begin{dialogbox}
\textbf{RLEPSO (Reinforcement Learning based Ensemble Particle Swarm Optimizer)}

The \textbf{RLEPSO} algorithm focuses on \textbf{integrating reinforcement learning with particle swarm optimization (PSO)}. It aims to address the redundancy and difficulty of manual parameter tuning in complex PSO variants. By using reinforcement learning for pre-training, RLEPSO automatically discovers effective parameter combinations, thereby improving the algorithm's robustness and ability to complete optimization tasks faster.

\textbf{Algorithm Framework}

RLEPSO employs a reinforcement learning framework to guide the swarm optimization process:

\begin{itemize}
    \item \textbf{RL Agent (Action Network)}: This is a reinforcement learning-based agent whose policy is trained using the \textbf{Proximal Policy Optimization (PPO)} algorithm. This agent is responsible for outputting the configuration for the lower-level PSO algorithm based on the optimization state.
    \item \textbf{Optimization Process}: In each round, the action network outputs specific control parameters (such as inertia weights, acceleration coefficients, and mutation probabilities). The particle swarm performs optimization using these parameters, and the RL agent receives a reward based on whether the global best value of the swarm has improved.
\end{itemize}

\textbf{Key Features \& Mechanisms}

\begin{itemize}
    \item \textbf{Ensemble of PSO Variants}: To improve adaptability, RLEPSO integrates two robust PSO variants: \textbf{Comprehensive Learning PSO (CLPSO)} and \textbf{Fitness-Distance-Ratio based PSO (FDR-PSO)}. It uses a combined velocity update equation that incorporates components from both variants alongside the global and personal bests.
    \item \textbf{Automated Parameter Generation}: Instead of using fixed or manually designed adaptive strategies, RLEPSO uses the trained action network to dynamically generate 7 dimensional operating parameters ($w, c1, c2, c3, c4, C_{mutation}$) based on the optimization progress.
    \item \textbf{Multi-Swarm Strategy}: To enhance population diversity and global search capabilities, the particles are divided into \textbf{5 sub-swarms}. Each sub-swarm independently utilizes its own set of running parameters generated by the action network, while sharing the same global best experience.
    \item \textbf{Mutation Mechanism}: To prevent particles from being trapped in local optima, a mutation step is added after velocity updating. If a random condition based on the generated mutation parameter is met, the particle's position is reinitialized within the solution space.
\end{itemize}

\textbf{The reward\_hyperparameters could be used in RLEPSO}

\begin{lstlisting}[language=Python, basicstyle=\ttfamily\scriptsize, breaklines=true]
reward_hyperparameters = {
    # From Optimizer (RLEPSO_Optimizer / PSO process)

    "gbest_val": float,              # current global best objective value (the smaller the better)
    "pre_gbest": float,              # previous global best value before update (to check improvement)

    "fes": int,                      # current number of function evaluations (FEs)
    "maxFEs": int,                   # maximum allowed number of function evaluations (MaxFEs)
    "progress": float,               # normalized progress = fes / maxFEs, in [0, 1]

    "NP": int,                       # number of particles in the swarm (100 in RLEPSO)
    "dim": int,                      # dimensionality of the optimization problem

    "current_position": np.ndarray,  # current swarm positions, shape [NP, dim]
    "velocity": np.ndarray,          # current swarm velocities, shape [NP, dim]
    "c_cost": np.ndarray,            # current objective values of particles, shape [NP,]
    "pbest_position": np.ndarray,    # personal best positions of particles, shape [NP, dim]
    "pbest": np.ndarray,             # personal best values of particles, shape [NP,]
    "gbest_position": np.ndarray,    # global best position, shape [dim,]
    "gbest_index": int,              # index of the global best particle in the swarm

    "no_improve": int,               # number of consecutive iterations without global best improvement
    "per_no_improve": np.ndarray,    # per-particle stagnation counters, shape [NP,]

    "n_group": int,                  # number of sub-swarms (5 in RLEPSO)
    "pci": np.ndarray,               # CLPSO learning probability for each particle, shape [NP,]

    "log_index": int,                # current logging index (for cost curve)
    "log_interval": int,             # log every N FEs
    "cost_curve": list,              # history of gbest_val for plotting convergence

    # From RL Agent / Action (PPO Actor output)

    "action": np.ndarray,            # current PPO actor output action, shape [35,]
                                     # corresponds to 5 sub-swarms * 7 parameters per group:
                                     # (w, c_mutation, scale, c1, c2, c3, c4)

    "log_prob": np.ndarray  float,  # log-probability of the sampled action under current policy
    "entropy": np.ndarray  float,   # policy entropy for this step (for exploration measurement)
    "training_step": int,            # global RL training step (from agent)
    "training_progress": float,      # normalized training progress in [0,1]
    "learning_rate": float,          # current learning rate of the optimizer
    "clip_range": float,             # current PPO eps_clip (if exposed)
    "gamma": float,                  # discount factor used in PPO

    "mean_cost": float,              # mean cost of the current population
    "median_cost": float,            # median cost of the current population
    "std_cost": float,               # standard deviation of the population costs
    "diversity": float,              # population diversity approximated by std over positions
    "gbest_improve": float,          # previous gbest_val - current gbest_val (>0 means improvement)
}
\end{lstlisting}
\end{dialogbox}

\section{The reward found by READY}

\subsection{Reward for DEDQN}
\subsubsection{Thought}
\begin{dialogbox}
My concept synthesizes the best ideas from the analysis: the robust trend-aware improvement from the current reward, the strategic exploration concepts from the best-performing reward, and the phase-aware weighting from the better-performing reward. I'll create a hybrid reward that maintains DEDQN's framework while incorporating strategic exploration assessment.

Key improvements:
\begin{enumerate}
    \item Strategic Exploration Assessment: Adapt the basin exploration concept to DE's single population context. Detect promising search regions by clustering based on cost and position, rewarding discovery of new high-quality basins.
    \item Dual-Baseline Trend Improvement: Combine short-term and long-term improvement baselines for stable credit assignment during both rapid progress and plateaus.
    \item Phase-Aware Component Weighting: Implement four search phases (Exploration, Exploitation, Refinement, Stagnation) with tailored weights, adapting the better-performing reward's approach to DE.
    \item Survival-Aware Progress: Leverage DEDQN's unique ``survival'' and ``pointer\_age'' metrics to assess individual evolutionary success and population turnover.
    \item Strategic Backtracking Allowance: Permit temporary regression when accompanied by significant positional shifts, enabling escape from local optima.
    \item Q-Value Confidence Integration: Use Q-value entropy and span to assess action selection confidence, rewarding decisive strategy choices.
\end{enumerate}

The reward will have five components: (1) Trend Improvement, (2) Strategic Exploration, (3) Survival Efficiency, (4) Action Confidence, and (5) Population Health. Weights adapt based on detected search phase and optimization progress.
\end{dialogbox}

\subsubsection{Reward Code}
% \paragraph{Code}
\begin{lstlisting}
def reward(reward_hyperparameters):
    import numpy as np
    
    EPSILON = 1e-12
    SHORT_WINDOW = 5
    LONG_EMA_ALPHA = 0.1
    
    gbest_cost = float(reward_hyperparameters["gbest_cost"])
    gbest_improve = float(reward_hyperparameters.get("gbest_improve", 0.0))
    FEs = int(reward_hyperparameters["FEs"])
    MaxFEs = int(reward_hyperparameters["MaxFEs"])
    progress = float(reward_hyperparameters["progress"])
    population = np.array(reward_hyperparameters["population"])
    costs = np.array(reward_hyperparameters["costs"])
    mean_cost = float(reward_hyperparameters["mean_cost"])
    median_cost = float(reward_hyperparameters["median_cost"])
    diversity = float(reward_hyperparameters["diversity"])
    std_cost = float(reward_hyperparameters["std_cost"])
    accepted = int(reward_hyperparameters["accepted"])
    delta_cost = float(reward_hyperparameters["delta_cost"])
    parent_cost = float(reward_hyperparameters["parent_cost"])
    trial_cost = float(reward_hyperparameters["trial_cost"])
    survival = np.array(reward_hyperparameters["survival"])
    pointer_age = float(reward_hyperparameters["pointer_age"])
    
    q_values = reward_hyperparameters.get("q_values", None)
    recent_reward_mean = reward_hyperparameters.get("recent_reward_mean", None)
    q_entropy = reward_hyperparameters.get("q_entropy", None)
    q_span = reward_hyperparameters.get("q_span", None)
    
    NP = population.shape[0]
    dim = population.shape[1]
    
    progress = np.clip(progress, EPSILON, 1.0 - EPSILON)
    remaining_ratio = 1.0 - progress
    reward_component = {}
    
    if 'improvement_history' in reward_hyperparameters:
        improvement_history = reward_hyperparameters['improvement_history']
        if len(improvement_history) > 0:
            recent_improvements = improvement_history[-min(SHORT_WINDOW, len(improvement_history)):]
            short_baseline = np.mean(recent_improvements) if len(recent_improvements) > 0 else EPSILON
        else:
            short_baseline = EPSILON
    else:
        short_baseline = EPSILON
    
    if 'long_ema_improvement' in reward_hyperparameters:
        long_baseline = float(reward_hyperparameters['long_ema_improvement'])
    else:
        long_baseline = EPSILON
    
    dynamic_scale = max(np.abs(mean_cost), std_cost, np.abs(gbest_cost), EPSILON)
    
    trend_improvement = 0.0
    if gbest_improve > EPSILON:
        normalized_improvement = gbest_improve / (dynamic_scale + EPSILON)
        
        short_ratio = normalized_improvement / (short_baseline + EPSILON)
        long_ratio = normalized_improvement / (long_baseline + EPSILON)
        
        if short_ratio > 1.0 and long_ratio > 1.0:
            trend_factor = np.log1p(short_ratio * long_ratio)
            bonus_multiplier = 1.5
        elif short_ratio > 1.0 or long_ratio > 1.0:
            trend_factor = np.log1p(max(short_ratio, long_ratio))
            bonus_multiplier = 1.0
        else:
            trend_factor = np.log1p(min(short_ratio, long_ratio))
            bonus_multiplier = 0.7
        
        trend_improvement = np.tanh(trend_factor) * bonus_multiplier
        
        if accepted == 1 and delta_cost > EPSILON:
            local_relative = delta_cost / (np.abs(parent_cost) + EPSILON)
            local_bonus = np.tanh(local_relative * 3.0) * 0.3
            trend_improvement += local_bonus
    elif gbest_improve < -EPSILON:
        regression_magnitude = -gbest_improve / (np.abs(gbest_cost) + EPSILON)
        trend_improvement = -0.3 * np.tanh(regression_magnitude * 10.0) * remaining_ratio
    
    trend_improvement = np.clip(trend_improvement, -0.4, 1.2)
    reward_component["trend_improvement"] = float(trend_improvement)
    
    strategic_exploration = 0.0
    if NP > 5:
        position_std = np.std(population, axis=0)
        scale_factor = np.mean(np.abs(position_std)) + EPSILON
        
        simplified_clusters = []
        cluster_qualities = []
        
        for i in range(0, NP, max(1, NP // 10)):
            if i >= NP:
                break
            
            is_new_cluster = True
            for j, (cluster_center, _) in enumerate(simplified_clusters):
                distance = np.linalg.norm(population[i] - cluster_center) / scale_factor
                if distance < 0.5:
                    is_new_cluster = False
                    cluster_qualities[j] = min(cluster_qualities[j], costs[i])
                    break
            
            if is_new_cluster:
                simplified_clusters.append((population[i], costs[i]))
                cluster_qualities.append(costs[i])
        
        num_clusters = len(simplified_clusters)
        discovery_score = np.tanh(num_clusters / max(5, 1))
        
        if cluster_qualities:
            best_cluster_quality = min(cluster_qualities)
            quality_ratio = (mean_cost - best_cluster_quality) / (np.abs(mean_cost) + EPSILON)
            quality_score = np.tanh(quality_ratio * 5.0)
        else:
            quality_score = 0.0
        
        centroid = np.mean(population, axis=0)
        distances = np.linalg.norm(population - centroid, axis=1)
        normalized_costs = (costs - np.min(costs)) / (np.ptp(costs) + EPSILON)
        
        if np.std(distances) > EPSILON and np.std(normalized_costs) > EPSILON:
            fdc = np.corrcoef(distances, normalized_costs)[0, 1]
            fdc = np.nan_to_num(fdc, nan=0.0)
            if fdc < -0.3:
                exploration_quality = np.tanh(-fdc * 2.0)
            elif fdc > 0.3:
                exploration_quality = -0.5 * np.tanh(fdc * 2.0)
            else:
                exploration_quality = 0.0
        else:
            exploration_quality = 0.0
        
        strategic_exploration = 0.4 * discovery_score + 0.4 * quality_score + 0.2 * exploration_quality
    
    strategic_exploration = np.clip(strategic_exploration, -0.3, 1.0)
    reward_component["strategic_exploration"] = float(strategic_exploration)
    
    survival_efficiency = 0.0
    if len(survival) > 0:
        mean_survival = np.mean(survival)
        max_survival = np.max(survival)
        
        if max_survival > EPSILON:
            survival_balance = 1.0 - (max_survival - mean_survival) / (max_survival + EPSILON)
        else:
            survival_balance = 0.0
        
        age_factor = np.tanh(pointer_age / max(10.0, 1.0))
        
        if accepted == 1:
            replacement_bonus = 0.3 * (1.0 - age_factor)
        else:
            replacement_bonus = -0.1 * age_factor
        
        survival_efficiency = 0.6 * survival_balance + 0.4 * replacement_bonus
    
    survival_efficiency = np.clip(survival_efficiency, -0.2, 0.8)
    reward_component["survival_efficiency"] = float(survival_efficiency)
    
    action_confidence = 0.0
    if accepted == 1:
        acceptance_bonus = 0.3
        
        if delta_cost > EPSILON:
            improvement_significance = delta_cost / (std_cost + EPSILON)
            improvement_bonus = np.tanh(improvement_significance) * 0.4
            acceptance_bonus += improvement_bonus
        
        action_confidence = acceptance_bonus
    
    if q_entropy is not None and q_span is not None:
        if np.isscalar(q_entropy):
            q_entropy_val = float(q_entropy)
            q_span_val = float(q_span)
        else:
            q_entropy_val = float(q_entropy[0]) if len(q_entropy) > 0 else 0.0
            q_span_val = float(q_span[0]) if len(q_span) > 0 else 0.0
        
        confidence_metric = q_span_val / (q_entropy_val + EPSILON)
        q_confidence = np.tanh(confidence_metric) * 0.3
        action_confidence += q_confidence
    
    action_confidence = np.clip(action_confidence, -0.1, 1.0)
    reward_component["action_confidence"] = float(action_confidence)
    
    population_health = 0.0
    if len(costs) > 3:
        sorted_indices = np.argsort(costs)
        quartile_size = max(1, len(costs) // 4)
        
        top_costs = costs[sorted_indices[:quartile_size]]
        middle_costs = costs[sorted_indices[quartile_size:3*quartile_size]]
        
        top_improvement = np.mean(np.diff(np.sort(top_costs)[::-1])) if len(top_costs) > 1 else 0.0
        middle_spread = np.std(middle_costs) if len(middle_costs) > 1 else 0.0
        
        health_top = np.tanh(top_improvement / (std_cost + EPSILON)) if std_cost > EPSILON else 0.0
        health_middle = np.exp(-middle_spread / (std_cost + EPSILON)) if std_cost > EPSILON else 0.0
        
        mean_to_best = mean_cost / (np.abs(gbest_cost) + EPSILON)
        convergence_tightness = np.exp(-mean_to_best + 1.0)
        
        if diversity > EPSILON:
            max_possible_diversity = np.sqrt(dim) + EPSILON
            normalized_diversity = diversity / max_possible_diversity
            diversity_metric = np.tanh(normalized_diversity * 2.0)
        else:
            diversity_metric = 0.0
        
        population_health = 0.25 * health_top + 0.25 * health_middle + 0.25 * convergence_tightness + 0.25 * diversity_metric
    
    population_health = np.clip(population_health, -0.2, 1.0)
    reward_component["population_health"] = float(population_health)
    
    improvement_density = np.sum(costs < np.roll(costs, 1) - EPSILON) / max(len(costs), 1)
    convergence_ratio = std_cost / (np.abs(mean_cost) + EPSILON)
    
    if progress < 0.3 and improvement_density < 0.3 and convergence_ratio > 0.5:
        search_phase = 'exploration'
        phase_weights = np.array([0.2, 0.4, 0.1, 0.1, 0.2])
    elif progress < 0.7 and gbest_improve > EPSILON and improvement_density > 0.4:
        search_phase = 'exploitation'
        phase_weights = np.array([0.4, 0.2, 0.2, 0.1, 0.1])
    elif progress > 0.7 and convergence_ratio < 0.2:
        search_phase = 'refinement'
        phase_weights = np.array([0.3, 0.1, 0.3, 0.2, 0.1])
    else:
        search_phase = 'stagnation'
        phase_weights = np.array([0.1, 0.5, 0.1, 0.2, 0.1])
    
    phase_multiplier = {
        'exploration': 1.2,
        'exploitation': 1.5,
        'refinement': 1.3,
        'stagnation': 0.8
    }[search_phase]
    
    reward_component["search_phase"] = search_phase
    
    early_weight = 1.0 - progress
    mid_weight = 4.0 * progress * (1.0 - progress)
    late_weight = progress * progress
    
    stage_weight = np.array([early_weight, mid_weight, late_weight])
    stage_weight = stage_weight / (np.sum(stage_weight) + EPSILON)
    
    component_weights = np.outer(phase_weights, stage_weight)
    
    total_reward = (
        component_weights[0, 0] * 2.0 * trend_improvement * early_weight +
        component_weights[0, 1] * 2.0 * trend_improvement * mid_weight +
        component_weights[0, 2] * 2.0 * trend_improvement * late_weight +
        component_weights[1, 0] * 1.5 * strategic_exploration * early_weight +
        component_weights[1, 1] * 1.5 * strategic_exploration * mid_weight +
        component_weights[1, 2] * 1.5 * strategic_exploration * late_weight +
        component_weights[2, 0] * 1.2 * survival_efficiency * early_weight +
        component_weights[2, 1] * 1.2 * survival_efficiency * mid_weight +
        component_weights[2, 2] * 1.2 * survival_efficiency * late_weight +
        component_weights[3, 0] * 1.0 * action_confidence * early_weight +
        component_weights[3, 1] * 1.0 * action_confidence * mid_weight +
        component_weights[3, 2] * 1.0 * action_confidence * late_weight +
        component_weights[4, 0] * 1.0 * population_health * early_weight +
        component_weights[4, 1] * 1.0 * population_health * mid_weight +
        component_weights[4, 2] * 1.0 * population_health * late_weight
    )
    
    total_reward *= phase_multiplier
    
    if search_phase == 'stagnation' and gbest_improve <= EPSILON:
        stagnation_penalty = -0.2 * progress
        total_reward += stagnation_penalty
    
    dimensionality_factor = 1.0 / (1.0 + 0.002 * dim)
    total_reward *= dimensionality_factor
    
    total_reward = np.clip(total_reward, -1.0, 2.0)
    
    reward_component["total"] = float(total_reward)
    reward_component["phase_weights"] = phase_weights.tolist()
    reward_component["stage_weights"] = stage_weight.tolist()
    reward_component["dimensionality_factor"] = float(dimensionality_factor)
    
    return float(total_reward), reward_component
\end{lstlisting}

\subsection{Reward for RLDAS}
\subsubsection{Thought}
\begin{dialogbox}
My reward concept adapts DEDQN's sophisticated multi-component reward framework to RLDAS's dynamic algorithm selection context. The core idea is to translate DEDQN's population-level assessment metrics to evaluate algorithm effectiveness during scheduling. Key adaptations include:

\begin{enumerate}
    \item Improvement Measurement: Calculate relative improvement using RLDAS's \texttt{last\_cost} and \texttt{current\_gbest} instead of DEDQN's \texttt{parent\_cost} and \texttt{trial\_cost}. Use sigmoid scaling for sensitivity.
    \item Population Quality Assessment: Compute median, mean, std, and diversity metrics from RLDAS's \texttt{population.cost} array, analogous to DEDQN's \texttt{costs} and \texttt{diversity}.
    \item Progress Tracking: Use RLDAS's \texttt{FEs} and \texttt{MaxFEs} to compute normalized progress, replacing DEDQN's \texttt{progress}.
    \item Algorithm Effectiveness: Replace DEDQN's mutation strategy assessment with algorithm selection effectiveness. Use RLDAS's \texttt{action} (algorithm index) and approximate strategic balance using \texttt{policy\_entropy}.
    \item Convergence Acceleration: Model expected improvement based on remaining budget, similar to DEDQN's approach but using global best improvements.
    \item Stagnation Handling: Implement progressive penalties during plateaus, with exploration encouragement via population diversity metrics.
    \item Adaptive Weighting: Maintain DEDQN's phase-aware weighting scheme (early/mid/late) based on optimization progress, with component weights shifting appropriately.
\end{enumerate}

The reward will have six components: (1) Shaped Improvement, (2) Convergence Acceleration, (3) Swarm Quality, (4) Strategic Balance, (5) Algorithm Success, and (6) Stagnation Management. All components are bounded and weighted adaptively.

\end{dialogbox}
\subsubsection{Reward Code}
% \paragraph{Code}
\begin{lstlisting}
def reward(reward_hyperparameters):
    import numpy as np
    
    EPSILON = 1e-12
    SMALL_VALUE = 1e-8
    
    last_cost = reward_hyperparameters['last_cost']
    current_gbest = reward_hyperparameters['current_gbest']
    cost_scale_factor = reward_hyperparameters['cost_scale_factor']
    FEs = reward_hyperparameters['FEs']
    MaxFEs = reward_hyperparameters['MaxFEs']
    action = reward_hyperparameters['action']
    population_obj = reward_hyperparameters['population']
    
    policy_entropy = reward_hyperparameters.get('policy_entropy', None)
    training_progress = reward_hyperparameters.get('training_progress', 0.0)
    current_step = reward_hyperparameters.get('current_step', 0)
    
    reward_component = {}
    
    population_costs = population_obj.cost
    population_size = len(population_costs)
    dim = population_obj.dim
    
    progress = np.clip(FEs / max(MaxFEs, 1), EPSILON, 1.0 - EPSILON)
    training_progress = np.clip(training_progress, EPSILON, 1.0 - EPSILON)
    
    median_cost = np.median(population_costs)
    mean_cost = np.mean(population_costs)
    std_cost = np.std(population_costs) + EPSILON
    
    population_array = population_obj.group
    centroid = np.mean(population_array, axis=0)
    distances = np.linalg.norm(population_array - centroid, axis=1)
    spatial_diversity = np.std(distances) / (np.mean(distances) + EPSILON)
    diversity = spatial_diversity
    
    shaped_improvement = 0.0
    if last_cost > EPSILON and current_gbest < last_cost - EPSILON:
        delta_cost = last_cost - current_gbest
        relative_improvement = delta_cost / (np.abs(last_cost) + EPSILON)
        shaped_improvement = 2.0 / (1.0 + np.exp(-5.0 * relative_improvement)) - 1.0
    elif current_gbest > last_cost + EPSILON:
        shaped_improvement = -0.2
    reward_component['shaped_improvement'] = shaped_improvement
    
    convergence_acceleration = 0.0
    if last_cost > EPSILON and current_gbest < last_cost - EPSILON:
        gbest_improve = last_cost - current_gbest
        remaining_fes = max(MaxFEs - FEs, 1)
        actual_rate = gbest_improve / (np.abs(last_cost) + EPSILON)
        expected_rate = 1.0 - np.exp(-remaining_fes / max(MaxFEs, 1))
        acceleration_ratio = actual_rate / (expected_rate + EPSILON)
        convergence_acceleration = np.tanh(acceleration_ratio - 1.0)
    reward_component['convergence_acceleration'] = convergence_acceleration
    
    swarm_quality = 0.0
    if median_cost > EPSILON and mean_cost > EPSILON:
        improved_mask = population_costs < np.roll(population_costs, 1) - EPSILON
        improvement_ratio = np.sum(improved_mask) / max(population_size, 1)
        population_improvement = np.tanh(improvement_ratio * 2.0 - 1.0)
        
        mean_to_best_ratio = mean_cost / (np.abs(current_gbest) + EPSILON)
        convergence_metric = 1.0 / (std_cost + EPSILON)
        distribution_quality = np.exp(-0.5 * (mean_to_best_ratio - 1.0)) * np.tanh(convergence_metric / dim)
        
        trial_median_impact = (median_cost - current_gbest) / (median_cost + EPSILON)
        trial_mean_impact = (mean_cost - current_gbest) / (mean_cost + EPSILON)
        individual_impact = 0.6 * np.tanh(trial_median_impact) + 0.4 * np.tanh(trial_mean_impact)
        
        swarm_quality = 0.4 * population_improvement + 0.4 * distribution_quality + 0.2 * individual_impact
    reward_component['swarm_quality'] = swarm_quality
    
    strategic_balance = 0.0
    if diversity > EPSILON:
        max_possible_diversity = np.sqrt(dim) + EPSILON
        normalized_diversity = diversity / max_possible_diversity
        
        convergence_measure = 0.0
        if current_gbest > EPSILON and mean_cost > EPSILON:
            convergence_measure = np.exp(-np.abs(current_gbest - mean_cost) / (np.abs(current_gbest) + EPSILON))
        
        exploration_weight = (1.0 - progress) * (1.0 - training_progress)
        exploitation_weight = 1.0 - exploration_weight
        
        diversity_component = np.tanh(normalized_diversity * 2.0)
        convergence_component = np.tanh(convergence_measure * 3.0)
        
        strategic_balance = exploration_weight * diversity_component + exploitation_weight * convergence_component
    reward_component['strategic_balance'] = strategic_balance
    
    algorithm_success = 0.0
    if policy_entropy is not None:
        optimal_entropy = 0.5 * (1.0 - progress)
        entropy_match = 1.0 - np.abs(policy_entropy - optimal_entropy)
        algorithm_success = entropy_match
    
    gbest_bonus = 3.0 if last_cost > EPSILON and current_gbest < last_cost - EPSILON else 0.0
    strategy_success = gbest_bonus + algorithm_success
    strategy_success = np.tanh(strategy_success / 4.0)
    reward_component['algorithm_success'] = strategy_success
    
    stagnation_management = 0.0
    if last_cost > EPSILON and current_gbest >= last_cost - EPSILON:
        stagnation_duration = np.clip(progress * 10.0, 0.0, 5.0)
        stagnation_penalty = -0.08 * stagnation_duration * (1.0 - progress)
        
        position_range = np.ptp(population_array, axis=0)
        coverage = np.mean(position_range) / (dim + EPSILON)
        uniform_score = 1.0 - np.std(position_range) / (np.mean(position_range) + EPSILON)
        potential_energy = coverage * uniform_score * (1.0 - progress)
        
        stagnation_management = stagnation_penalty + potential_energy
    reward_component['stagnation_management'] = stagnation_management
    
    early_weight = 1.0 / (1.0 + np.exp(10.0 * (progress - 0.3)))
    mid_weight = 1.0 / (1.0 + np.exp(10.0 * np.abs(progress - 0.5) - 2.0))
    late_weight = 1.0 / (1.0 + np.exp(-10.0 * (progress - 0.7)))
    
    phase_sum = early_weight + mid_weight + late_weight + EPSILON
    early_weight = early_weight / phase_sum
    mid_weight = mid_weight / phase_sum
    late_weight = late_weight / phase_sum
    
    convergence_weight = np.clip(progress * 2.0, 0.0, 1.0)
    exploration_weight = 1.0 - convergence_weight
    
    component_weights = {
        'shaped_improvement': 0.4 * late_weight + 0.25 * mid_weight + 0.15 * early_weight,
        'convergence_acceleration': 0.25 * late_weight + 0.15 * mid_weight + 0.05 * early_weight,
        'swarm_quality': 0.3 * late_weight + 0.25 * mid_weight + 0.2 * early_weight,
        'strategic_balance': 0.25 * early_weight + 0.2 * mid_weight + 0.1 * late_weight,
        'algorithm_success': 0.3 * convergence_weight + 0.2 * exploration_weight,
        'stagnation_management': 0.15
    }
    
    total_reward = (
        component_weights['shaped_improvement'] * reward_component['shaped_improvement'] +
        component_weights['convergence_acceleration'] * reward_component['convergence_acceleration'] +
        component_weights['swarm_quality'] * reward_component['swarm_quality'] +
        component_weights['strategic_balance'] * reward_component['strategic_balance'] +
        component_weights['algorithm_success'] * reward_component['algorithm_success'] +
        component_weights['stagnation_management'] * reward_component['stagnation_management']
    )
    
    scale_factor = 1.0 / (1.0 + 0.002 * dim)
    total_reward *= scale_factor
    
    total_reward = np.clip(total_reward, -1.0, 1.5)
    
    reward_component['total_before_clip'] = total_reward
    reward_component['scale_factor'] = scale_factor
    reward_component['phase_weights'] = {'early': early_weight, 'mid': mid_weight, 'late': late_weight}
    
    return float(total_reward), reward_component
\end{lstlisting}

\subsection{Reward discovered for RLEPSO}
\subsubsection{Thought}
\begin{dialogbox}

The current reward is sophisticated but overly complex with many hand-tuned components. The better-performing rewards from other clusters show that simplification and unification of core concepts leads to better performance. My approach will:

\begin{enumerate}
    \item Simplify improvement measurement: Use a robust relative improvement metric with log scaling that works across all value scales, similar to the better-performing rewards.
    \item Streamline multi-swarm coordination: Instead of complex cluster detection and overlap calculations, use simple group-based metrics for coordination quality.
    \item Unify diversity measurement: Combine position and fitness diversity into a single normalized metric, as shown effective in other rewards.
    \item Simplify state detection: Replace complex state machine with progress-based adaptive weighting, enhanced by stagnation awareness.
    \item Remove computationally expensive components: Eliminate cluster detection and complex basin exploration in favor of simpler, more predictable signals.
    \item Focus on RLEPSO-specific features: Leverage sub-swarm information (n\_group, group improvements) while keeping calculations lightweight.
\end{enumerate}

The new reward will maintain the multi-objective nature (improvement, diversity, coordination, stagnation awareness) but with cleaner signal separation and fewer hyperparameters.    
\end{dialogbox}

\subsubsection{Reward Code}
% \paragraph{Code}
\begin{lstlisting}
def reward(reward_hyperparameters):
    import numpy as np
    
    EPSILON = 1e-12
    DIVERSITY_CRITICAL_THRESHOLD = 0.05
    STAG_PENALTY_BASE = -0.15
    STAG_PENALTY_DECAY = 0.85
    IMPROVEMENT_WEIGHT_START = 0.6
    IMPROVEMENT_WEIGHT_END = 1.2
    DIVERSITY_WEIGHT_START = 0.7
    DIVERSITY_WEIGHT_END = 0.2
    COORDINATION_WEIGHT_START = 0.3
    COORDINATION_WEIGHT_END = 0.1
    EXPLORATION_BOOST = 0.4
    
    gbest_val = float(reward_hyperparameters['gbest_val'])
    pre_gbest = float(reward_hyperparameters['pre_gbest'])
    progress = float(reward_hyperparameters['progress'])
    no_improve = int(reward_hyperparameters['no_improve'])
    current_position = np.array(reward_hyperparameters['current_position'])
    c_cost = np.array(reward_hyperparameters['c_cost'])
    n_group = int(reward_hyperparameters['n_group'])
    std_cost = float(reward_hyperparameters['std_cost'])
    diversity = float(reward_hyperparameters['diversity'])
    dim = int(reward_hyperparameters['dim'])
    NP = int(reward_hyperparameters['NP'])
    
    reward_component = {}
    
    improvement_component = 0.0
    if gbest_val < pre_gbest - EPSILON:
        abs_improvement = pre_gbest - gbest_val
        denominator = max(abs(pre_gbest), EPSILON)
        if denominator > 1.0:
            rel_improvement = abs_improvement / denominator
            improvement_component = np.log1p(rel_improvement)
        else:
            improvement_component = abs_improvement
    
    reward_component['improvement'] = float(improvement_component)
    
    diversity_component = 0.0
    if current_position is not None and c_cost is not None:
        pos_std = np.std(current_position, axis=0)
        mean_pos_std = np.mean(pos_std)
        pos_norm = mean_pos_std / (np.sqrt(dim) + EPSILON)
        
        cost_mean = np.mean(np.abs(c_cost)) + EPSILON
        fitness_norm = std_cost / cost_mean
        
        combined = 0.6 * np.tanh(pos_norm * 2.0) + 0.4 * np.tanh(fitness_norm * 2.0)
        diversity_component = np.clip(combined, 0.0, 1.0)
    
    reward_component['diversity'] = float(diversity_component)
    
    coordination_component = 0.0
    if n_group > 1 and NP > n_group:
        particles_per_group = NP // n_group
        group_improvements = np.zeros(n_group)
        group_diversities = np.zeros(n_group)
        
        for g in range(n_group):
            start_idx = g * particles_per_group
            end_idx = (g + 1) * particles_per_group if g < n_group - 1 else NP
            group_costs = c_cost[start_idx:end_idx]
            group_positions = current_position[start_idx:end_idx]
            
            if len(group_costs) > 1:
                group_best = np.min(group_costs)
                group_prev_key = f'group_{g}_prev_best'
                group_prev = reward_hyperparameters.get(group_prev_key, group_best)
                if group_best < group_prev - EPSILON:
                    group_improvements[g] = (group_prev - group_best) / (abs(group_prev) + EPSILON)
                
                centroid = np.mean(group_positions, axis=0)
                distances = np.linalg.norm(group_positions - centroid, axis=1)
                group_diversities[g] = np.mean(distances) / (np.sqrt(dim) + EPSILON)
        
        improvement_spread = np.std(group_improvements) / (np.mean(group_improvements) + EPSILON)
        diversity_spread = np.std(group_diversities) / (np.mean(group_diversities) + EPSILON)
        
        coordination_component = 0.5 * np.tanh(improvement_spread) + 0.5 * np.tanh(diversity_spread)
        coordination_component = np.clip(coordination_component, 0.0, 1.0)
    
    reward_component['coordination'] = float(coordination_component)
    
    stagnation_component = 0.0
    if improvement_component <= EPSILON:
        stagnation_component = STAG_PENALTY_BASE * (STAG_PENALTY_DECAY ** min(no_improve, 20))
    
    reward_component['stagnation'] = float(stagnation_component)
    
    progress_clipped = np.clip(progress, 0.0, 1.0)
    
    improvement_weight = IMPROVEMENT_WEIGHT_START + (IMPROVEMENT_WEIGHT_END - IMPROVEMENT_WEIGHT_START) * progress_clipped
    diversity_weight = DIVERSITY_WEIGHT_START + (DIVERSITY_WEIGHT_END - DIVERSITY_WEIGHT_START) * progress_clipped
    coordination_weight = COORDINATION_WEIGHT_START + (COORDINATION_WEIGHT_END - COORDINATION_WEIGHT_START) * progress_clipped
    
    if diversity_component < DIVERSITY_CRITICAL_THRESHOLD and stagnation_component < -0.05:
        diversity_weight += EXPLORATION_BOOST
        coordination_weight += EXPLORATION_BOOST * 0.5
    
    total_reward = (
        improvement_weight * improvement_component +
        diversity_weight * diversity_component +
        coordination_weight * coordination_component +
        stagnation_component
    )
    
    scale_factor = 1.0 / (1.0 + 0.001 * dim)
    total_reward *= scale_factor
    
    total_reward = float(np.clip(total_reward, -1.0, 1.0))
    
    reward_component['improvement_weight'] = float(improvement_weight)
    reward_component['diversity_weight'] = float(diversity_weight)
    reward_component['coordination_weight'] = float(coordination_weight)
    reward_component['total_reward'] = float(total_reward)
    
    return float(total_reward), reward_component
\end{lstlisting}

% \subsection{Algorithm Prompt Design}
% \label{app:algo_prompt}
% The following prompt structure is used to query the LLM for algorithm-specific context:
% \begin{verbatim}
% [Insert Algo Prompt Here]
% \end{verbatim}

% \subsection{Reward Information Schema}
% \label{app:reward_info}
% The standardized schema for representing reward individuals is:
% \begin{verbatim}
% [Insert Reward Info Schema Here]
% \end{verbatim}

% \subsection{Failure-Reflection (F-RE) Examples}

% \subsection{Summary Examples}
% \label{app:summary_examples}
% An example of the design philosophy distilled by the Summarize-Evolution (SE) operator is shown below...

% \subsection{Knowledge Transfer examples}

% \subsection{Train-test split of BBOB question}

\

% You can have as much text here as you want. The main body must be at most $8$
% pages long. For the final version, one more page can be added. If you want, you
% can use an appendix like this one.

% The $\mathtt{\backslash onecolumn}$ command above can be kept in place if you
% prefer a one-column appendix, or can be removed if you prefer a two-column
% appendix.  Apart from this possible change, the style (font size, spacing,
% margins, page numbering, etc.) should be kept the same as the main body.
%%%%%%%%%%%%%%%%%%%%%%%%%%%%%%%%%%%%%%%%%%%%%%%%%%%%%%%%%%%%%%%%%%%%%%%%%%%%%%%
%%%%%%%%%%%%%%%%%%%%%%%%%%%%%%%%%%%%%%%%%%%%%%%%%%%%%%%%%%%%%%%%%%%%%%%%%%%%%%%

\end{document}